\documentclass{article}

\usepackage{PRIMEarxiv}

\usepackage[utf8]{inputenc} 
\usepackage[T1]{fontenc}    
\usepackage{hyperref}       
\usepackage{url}            
\usepackage{booktabs}       
\usepackage{amsfonts}       
\usepackage{nicefrac}       
\usepackage{microtype}      
\usepackage{lipsum}
\usepackage{fancyhdr}       
\usepackage{graphicx}       
\graphicspath{{media/}}     
\usepackage{booktabs}
\usepackage{multirow}
\usepackage{array}
\usepackage{amsmath}
\usepackage[linesnumbered, ruled, vlined]{algorithm2e}
\usepackage{longtable}
\usepackage{caption}
\usepackage{float} 
\usepackage{array,tabularx}

\title{Inference-Time Optimization of Prompt Embeddings in Diffusion Models: A Comparison of sep-CMA-ES and Adam 
}

\author{
  Domício Pereira Neto\\
  University of Coimbra \\
  Coimbra, Portugal\\
  \texttt{dneto@dei.uc.pt} \\
   \And
  João Correia\\
  University of Coimbra \\
  Coimbra, Portugal\\
  \texttt{jncor@dei.uc.pt} \\
   \And
  Penousal Machado\\
  University of Coimbra \\
  Coimbra, Portugal\\
  \texttt{machado@dei.uc.pt} \\
}

\begin{document}
\maketitle

\begin{abstract}
Deep diffusion models have revolutionized image generation by producing high-quality outputs. However, achieving specific objectives with these models often requires costly adaptations such as fine-tuning, which can be resource-intensive and time-consuming. An alternative approach is inference-time control, which involves optimizing the prompt embeddings to guide the generation process without altering the model weights. We explore prompt-embedding search optimization for the Stable Diffusion XL Turbo model, comparing a gradient-free evolutionary approach, the Separable Covariance Matrix Adaptation Evolution Strategy (sep-CMA-ES), against the widely used gradient-based optimizer Adaptive Moment Estimation (Adam). Candidate images are evaluated by a weighted objective that combines LAION Aesthetic Predictor V2 and CLIPScore, enabling explicit trade-offs between aesthetic quality and prompt--image alignment. On 36 prompts sampled from Parti Prompts (P2) under three weight settings (aesthetics-only, balanced, alignment-only), sep-CMA-ES consistently achieves higher objective values than Adam. We additionally analyze divergence from the unoptimized baseline using cosine similarity and SSIM and report the compute and memory footprints. These results suggest that sep-CMA-ES is an effective inference-time optimizer for prompt-embedding search, improving aesthetics--alignment trade-offs and resource usage without model fine-tuning.
\end{abstract}


\section{Introduction}
Diffusion-based generative models have enabled high-fidelity image synthesis across multiple modalities; however, steering a frozen generator toward explicit objectives is still a challenging task without costly model adaptation (e.g., fine-tuning). Considering text-to-image generation, a single prompt can correspond to many plausible outputs because these models compress large-scale training data into high-dimensional latent representations \cite{Rombach2022}. In practice, standard prompting, that is, manually writing and testing prompts, explores only a small portion of the model's generative capacity, and achieving specific targets, such as improving aesthetics while preserving semantic faithfulness, can be difficult, particularly in settings where model internals are not accessible or controllable beyond the prompt interface~\cite{Li2025review}.

A lightweight alternative to fine-tuning or retraining is \emph{inference-time optimization} over the inputs that condition generation. Instead of updating model weights, one can search over continuous variables such as text-conditioning embeddings and select candidates using automatic evaluators. This casts controllable generation as an optimization problem: each update requires generating images and scoring them, and the resulting objective landscape is often highly non-convex, noisy, and expensive to evaluate. While gradient-based optimizers such as Adam (and variants such as AdamW) are the default choice for training and fine-tuning deep generative models~\cite{Kingma2014,loshchilov2019}, their use at inference time can be limited by weak or unstable gradients induced by stochastic sampling and multi-step denoising, restricted end-to-end differentiability when objectives depend on external or partially differentiable evaluators, and substantial memory overhead from storing intermediate activations when backpropagating through large, multi-component generative pipelines.

In this context, Evolutionary Machine Learning (EML) methods provide a natural fit. Evolutionary algorithms can optimize continuous variables using only function evaluations, maintain diverse candidate solutions, and explore solution spaces more broadly than purely local first-order methods. Therefore, they have been applied to image generation within evaluation-driven optimization settings, evolving prompts, latent variables, or embeddings under objectives related to quality, diversity, aesthetics, and alignment~\cite{Correia2024}. However, na\"ively applying powerful second-order evolutionary strategies in very high-dimensional spaces can be computationally prohibitive. This motivates scalable variants such as the \emph{Separable Covariance Matrix Adaptation Evolution Strategy} (sep-CMA-ES), which approximates the covariance matrix with a diagonal form, reducing time and memory complexity while retaining adaptive step-size control~\cite{ros2008}.

In this paper, we explore inference-time prompt-embedding optimization (i.e., optimization of the text encoder's continuous embeddings) for a frozen \emph{Stable Diffusion XL Turbo} generator~\cite{sauer2025}. We compare sep-CMA-ES~\cite{ros2008} against the widely used gradient-based optimizer \emph{Adam}~\cite{Kingma2014} on the same objective: a weighted combination of \emph{LAION Aesthetic Predictor V2}~\cite{Schuhmann2022} and \emph{CLIPScore}~\cite{hessel2021}. This formulation enables explicit trade-offs between aesthetic quality and prompt--image alignment. We evaluate both methods on 36 prompts sampled from \emph{Parti Prompts (P2)} under three weight settings (aesthetics-only, balanced, and alignment-only). Beyond the achieved objective values, we analyze divergence from the unoptimized baseline using cosine similarity and the Structural Similarity Index Measure (SSIM), and we report compute and memory footprints to characterize practical costs.

Across the three weight settings (aesthetics-only, balanced, and alignment-only), sep-CMA-ES achieves higher final fitness than Adam on the SDXL Turbo prompt-embedding task and attains the highest fitness on most prompts. Moreover, similarity-to-baseline analyses using cosine similarity and SSIM show that sep-CMA-ES typically departs further from the unoptimized generations than Adam, indicating a more exploratory search behavior under the same evaluation protocol.

The main contributions of this work are: (i) the \emph{Evolutionary Image Generation Optimization (EIGO)} engine, a reproducible optimization workflow of solution space search for diffusion models that integrates generation, automatic evaluation, and optimization using both evolutionary and gradient-based methods; (ii) a comparative analysis of sep-CMA-ES and Adam for inference-time prompt-embedding optimization under a multi-objective reward combining LAION Aesthetic Predictor V2 and CLIPScore~\cite{ros2008,Kingma2014,sauer2025,Schuhmann2022,hessel2021}; and (iii) an empirical study across three objective trade-offs, including similarity-to-baseline metrics (cosine similarity and SSIM) and compute and memory footprints to characterize exploration behavior and practical costs.

The remainder of the paper is organized as follows. Section~\ref{sec:related} reviews related work. Section~\ref{sec:approach} describes the methodology and the EIGO engine. Section~\ref{sec:setup} details the experimental setup, and Section~\ref{sec:results} presents results and analysis. Section~\ref{sec:conclusions} concludes and outlines future directions.

\section{Related Work}
\label{sec:related}

Deep generative models have rapidly progressed in their ability to synthesize high-quality images. Early work on conditional GANs showed that conditioning signals can enable controllable generation~\cite{mirza2014}, including strong spatial control through segmentation maps, as in SPADE~\cite{park2019}. Diffusion models have become the prevailing approach. These models generate images by iteratively denoising latent variables conditioned on text and/or other inputs. Transformer-based backbones, such as DiT~\cite{peebles2023} and distilled pipelines, allow high-fidelity generation with improved sampling efficiency, powering proprietary systems such as Google’s Imagen 3 \cite{imagenteamgoogle2024imagen3} and open models such as Stability AI’s Stable Diffusion 3 and FLUX \cite{esser2024,labs2025flux1kontextflowmatching}. 
Nevertheless, steering a \emph{frozen} generator toward explicit objectives is difficult because a single prompt can correspond to many plausible outputs depending on the model and parameters, and desirable regions of the generative space may be difficult to reach~\cite{Rombach2022,Li2025review}.

Model adaptation, namely fine-tuning, is a common method for improving controllability. For instance, DreamBooth fine-tunes a diffusion model to bind a unique identifier to a specific subject, enabling subject-driven generation~\cite{ruiz2022dreambooth}. In parallel, several methods have demonstrated that strong control signals can be injected at inference time without retraining the generator. Classifier-free guidance (CFG) improves sample quality by combining conditional and unconditional predictions and has become a standard inference-time control mechanism in diffusion pipelines~\cite{ho2022classifierfree}. SDEdit uses a diffusion prior for guided synthesis and editing by noising and denoising an input, balancing faithfulness and realism without task-specific training~\cite{meng2021sdedit}. These works demonstrate the feasibility of using inference-time strategies that treat conditioning inputs and sampling dynamics as primary levers for control.

Additionally, diffusion pipelines expose multiple intervention points (e.g., attention maps, latent trajectories, and text-conditioning embeddings) that can be manipulated to generate controllable outputs. Prompt-to-Prompt controls editing by intervening in cross-attention maps during diffusion, enabling localized and global edits driven by textual changes~\cite{hertz2022prompttoprompt}. DiffusionCLIP performs text-guided manipulation using diffusion dynamics and inversion, improving robustness over earlier GAN-inversion-based approaches~\cite{kim2021diffusionclip}. Although these methods target editing and control, they highlight that inference-time intervention is often practical, and multiple internal representations can be optimized.

Optimization of explicit objectives requires evaluation signals. Because human evaluation is costly, a large body of work has proposed automated measures of quality, diversity, faithfulness, and preference alignment~\cite{hu2023,yuval2023,liang2024,zhang2024}. ImageReward exemplifies learned preference modeling by training a reward model from expert comparisons and using reward feedback learning to improve diffusion models through fine-tuning~\cite{xu2023}. Open preference datasets and scorers have further improved the accessibility of preference-based evaluation. Pick-a-Pic collects large-scale user comparisons and trains PickScore, a CLIP-based preference predictor that correlates well with human rankings~\cite{kirstain2023pickapic}, whereas HPS v2 provides a large-scale preference benchmark and a tuned scoring model aimed at more reliable evaluation across distributions~\cite{wu2023hpsv2}. These works support inference-time optimization loops and show the importance of using complementary signals to mitigate evaluator biases and reduce the risk of optimizing toward artifacts of a single metric.

A line of work explores optimization at the level of text prompts, treating generation and evaluation as a loop in which candidate prompts are proposed and selected according to downstream scores~\cite{tran2023,hao2023,wong2023,wang2024,li2025}. MetaPrompter follows an interactive evolutionary approach in which users provide a meta-prompt and an Interactive Genetic Algorithm evolves concrete prompts, improving stylistic qualities while illustrating challenges in maintaining faithfulness and motivating automated evaluators~\cite{martins2023metaprompter}. Although prompt evolution is attractive due to its simplicity and compatibility with existing interfaces, the discrete nature of text can limit fine-grained control, motivating optimization in continuous spaces associated with generation.

To increase controllability, other studies directly perform searches in continuous spaces, including diffusion latents and text-conditioning embeddings~\cite{costa2023,haruka2023,clare2023,Wu2023,Yu2024}. ImageBreeder proposes an evolutionary inference-time framework that maintains populations of candidate images per prompt, scores them with ImageReward, and iteratively applies variation and selection operators in pixel or latent space~\cite{sobania2025}. Closest to our setting, Salvenmoser et al.\ optimized the prompt embedding vector search of a frozen SDXL Turbo model using a Genetic Algorithm and an aesthetic-only evaluator~\cite{salvenmoser2025}, demonstrating the promise of embedding-space search optimization but also highlighting the risk that single-metric objectives can drift away from prompt intent. This motivates objectives that explicitly combine complementary signals, such as aesthetics and prompt--image alignment.

Evolution strategies are well-suited for evaluation-driven optimization because they operate using only objective evaluations and can explore complex, noisy, and non-convex landscapes. CMA-ES is a canonical method for continuous optimization, adapting a sampling distribution to the landscape~\cite{hansen2001}, but scaling the standard CMA-ES to the dimensionality of prompt embeddings can be prohibitive. This motivates scalable variants, such as sep-CMA-ES, which uses a diagonal covariance approximation to achieve linear time and memory complexity while retaining adaptive step-size control~\cite{ros2008}. Building on the above literature, our work focuses on inference-time prompt-embedding search optimization and directly compares sep-CMA-ES with a widely used gradient-based optimizer (Adam) under a shared multiobjective that combines aesthetic quality and prompt--image alignment.

\section{Methodology}
\label{sec:approach}
This work compares two optimization algorithms, sep-CMA-ES and Adam, for inference-time prompt-embedding optimization in diffusion-based image generation. Using Stable Diffusion XL Turbo as the generative model, both algorithms were applied to optimize the text-conditioning embedding vector to improve image aesthetics and prompt--image semantic alignment, as measured by a weighted combination of LAION Aesthetic Predictor V2 and CLIPScore.

\subsection{EIGO}

To support the experimental workflow, we developed the Evolutionary Image Generation Optimization (EIGO) engine. EIGO is primarily designed for embedding optimization with CMA-ES and its variants, and Adam is included for comparison. EIGO is publicly available on GitHub \footnote{https://github.com/domiciopereiraneto/eigo} A walkthrough Jupyter Notebook is also provided along with the libraries developed for this work. 

\begin{figure}[htbp]
  \centering
  \includegraphics[width=\linewidth]{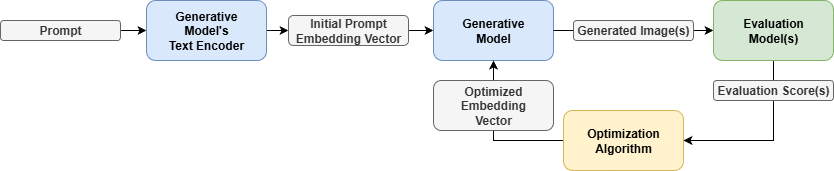}
  \caption{General structure and workflow of EIGO. The main components and their respective inputs and outputs are shown. The structure comprises two primary parts: the initialization phase and the optimization cycle. During the initialization phase, the prompt is converted into prompt embeddings, which are subsequently used in the optimization cycle. This cycle encompasses the following processes: (i) image generation, (ii) evaluation, and (iii) updating of the embeddings by the optimization algorithm.}
  \label{fig:framework}
\end{figure}

The architecture of the EIGO engine is illustrated in Figure~\ref{fig:framework}.

EIGO operates as follows: The text encoder of the generative model encodes a given prompt into an initial prompt-embedding vector. An initial image is then generated from the input embeddings without optimization and assessed using a weighted combination of metrics. The optimization algorithm updates the embedding vector to maximize the selected objective. This cycle between the generative model and the optimization algorithm continues until a specified number of iterations is completed or a time limit is reached. The final result is the best image obtained during the optimization run, determined by the highest weighted objective value.

EIGO is modular and can be coupled with different generative models, optimizers, and evaluation metrics. In this paper, we instantiate it with SDXL Turbo for generation, sep-CMA-ES and Adam for optimization, and LAION Aesthetic Predictor V2 plus CLIPScore for evaluation. The remainder of this section describes these components.

\subsection{SDXL Turbo}

Currently, there are numerous open-source image generation models, ranging from those with a few million parameters to hundreds of billions of parameters. Many state-of-the-art systems are diffusion-based and computationally expensive, including DeepFloyd-IF (three diffusion stages plus a large T5-family text encoder)~\cite{deepfloyd2023if}, DiT/PixArt-style diffusion transformers~\cite{chen2023}, and SDXL pipelines that use a base model and refiner with tens of sampling steps to reach $1024\times1024$ resolution~\cite{podell2023}. These are costly because they require many sequential denoising steps, multiple UNets or upsamplers, and large text encoders, which increase the FLOPs, memory, and latency while reducing the practical batch sizes. Therefore, we selected the well-established SDXL Turbo \cite{sauer2025},  a distilled variant of SDXL that produces high-quality images in one to four denoising steps, compared to the $\sim$50 steps typically used by standard SDXL.

\subsection{Image Evaluation}

In this work, we evaluate candidate images using a combination of aesthetic quality and prompt--image alignment. We summarize the two evaluation schemes below.


The LAION Aesthetic Predictor V2 is a lightweight regressor developed by the LAION community to estimate the human-perceived aesthetic quality of images on a scale of 1 to 10 \cite{Schuhmann2022}.  It was designed to help curate subsets of large web datasets (e.g., LAION-5B) and provide a fast automated score that is practical for inner-loop optimization. In our work, it contributes to the aesthetic component of the objective function.


To evaluate prompt--image alignment, we use CLIPScore, directly derived from OpenAI’s CLIP \cite{hessel2021}. The CLIP model produces an image embedding $f_I(x)$ and a text embedding $f_T(p)$; CLIPScore is their cosine similarity,

\begin{equation}
    \mathrm{CLIPScore}(x,p)=\frac{\langle f_I(x),f_T(p)\rangle}{\|f_I(x)\|\,\|f_T(p)\|},
\end{equation}

which provides an estimate of the semantic compatibility between prompt $p$ and image $x$.  Implementations may apply temperature scaling or normalization, but the core signal is this similarity, typically in $[-1,1]$.

CLIPScore is widely used for zero-shot classification, cross-modal retrieval, caption re-ranking, and evaluation or guidance in generative pipelines. It is fast, with only one forward pass through each encoder per sample; therefore, it scales to large sweeps and online selection. The known sensitivities include prompt wording, length, and dataset bias. In our experiments, we compute CLIPScore for each generated image and combine it with the LAION aesthetic score to form the objective optimized by both sep-CMA-ES and Adam.

\subsection{Optimization Algorithms}

The primary goal of this work is to assess evolutionary optimization for inference-time embedding search by comparing it with the current state-of-the-art gradient-based alternative.  Therefore, we compare sep-CMA-ES with the gradient-based Adam for the optimization of SDXL Turbo's prompt embeddings, which are presented in detail below.


CMA-ES is a powerful method for continuous optimization, but its standard formulation does not scale well to very high-dimensional problems. 
Standard CMA-ES samples candidates from a Gaussian $\mathcal N(m,\sigma^2 C)$ and adapts the covariance matrix $C$ using elite samples, with a time and memory complexity $O(d^2)$ for dimension $d$. Considering that the embedding space of deep generative models may reach tens of thousands of dimensions, applying CMA-ES becomes infeasible. Separable CMA-ES (sep-CMA-ES) addresses this by constraining $C$ to be diagonal and updating only coordinate-wise variances~\cite{ros2008}. This reduces the memory and time to $O(d)$ at the cost of ignoring cross-coordinate correlations. Assuming this compromise, we employed sep-CMA-ES to maximize the weighted sum of aesthetic quality and prompt alignment by optimizing the prompt embedding vector:

\noindent Let:
\begin{itemize}
    \item $\mathbf{z}\in\mathbb{R}^d$: prompt-embedding vector to be optimized;
    \item $p$: fixed text prompt;
    \item $G(\mathbf{z})$: generative model producing image $\mathbf{x}$;
    \item $S_{\text{aest}}(\mathbf{x})\in[1,10]$, $S_{\text{clip}}(\mathbf{x},p)\in[-1,1]$;
    \item $\hat S_{\text{aest}}(\mathbf{x})=\mathrm{norm}_a\!\left(S_{\text{aest}}(\mathbf{x})\right)\in[0,1]$;
    \item $\hat S_{\text{clip}}(\mathbf{x},p)=\mathrm{norm}_c\!\left(S_{\text{clip}}(\mathbf{x},p)\right)\in[0,1]$;
    \item $a,b\ge0$ (optionally $a+b=1$): metric weights.
\end{itemize}
Fitness is defined as
\begin{equation}
    F(\mathbf{z}) \;=\; a\,\hat S_{\text{aest}}\!\big(G(\mathbf{z})\big)\;+\; b\,\hat S_{\text{clip}}\!\big(G(\mathbf{z}),p\big),
    \label{eq:fitness}
\end{equation}
and the goal is:
\begin{equation}
    \mathbf{z}^* \;=\; \arg\max_{\mathbf{z}} \; F(\mathbf{z}).
    \label{eq:objective}
\end{equation}

Adam is a popular used optimizer in that iteratively updates parameters to minimize a loss function~\cite{Kingma2014}. It combines momentum-like first-moment estimates with adaptive learning rates based on second-moment estimates of the gradients, thus combining ideas from Momentum and RMSProp.
Adam updates both the gradients (first moment) and their squared values (second moment) using two moving averages, one for each iteration through an exponential decay. Subsequently, the averages are changed to account for bias, thus stabilizing early training updates. This method is usually regarded as computationally efficient and flexible for sparse and large-scale data problems. As it improves convergence and performance in complex, high-dimensional environments, Adam is extensively used to train neural networks, including those in the fields of Computer Vision, Natural Language Processing (NLP), and generative models.


Using Adam to optimize text embeddings can be effective because of its adaptive learning rate and effectiveness in high-dimensional optimization problems, where nonlinear interactions predominate. Therefore, in principle, Adam may provide fine-grained adjustments to achieve the desired aesthetic and image--prompt alignment optimization.  Nevertheless, it requires a differentiable end-to-end computation graph; in EIGO, this entails implementing a gradient-tracked evaluation path and an optimizer interface that reliably propagates gradients back to the embedding vector.


Consequently, we have the following loss function definition:

\begin{equation}
\mathcal{L}(\mathbf{z}) \;=\; 1-F(\mathbf{z})
\end{equation}

This minimizes the negative of the fitness function (Eq. \ref{eq:fitness}), setting the loss function between a maximum of 1 and a minimum of 0. All model weights are frozen; gradients flow only to $\mathbf{z}$.

\section{Experimental Setup}
\label{sec:setup}

As the guiding element of this comparison study, we chose the Parti Prompts (P2) dataset, which contains over 1600 prompts divided into 12 categories: Abstract, Vehicles, Illustrations, Arts, World Knowledge, People, Animals, Artifacts, Food \& Beverage, Produce \& Plants, Outdoor Scenes, and Indoor Scenes. Since running the optimization framework on the full dataset would require several thousand GPU hours, we randomly selected a smaller subset of 36 prompts (three per category). The selected prompts represented the following distribution of challenge types: Basic challenges (8), Fine-grained Detail (7), Simple Detail (5), Complex (5), Style \& Format (4), Imagination (2), Writing \& Symbols (2), Quantity (2), and Linguistic Structures (1).

The experiments consisted of running the optimization algorithms for each of the 36 prompts for 1000 seconds. The parameters of both algorithms were manually tuned and are detailed in Table~\ref{tab:parameters}, organized by optimizer; ``All'' denotes parameters shared across all experiments.

\begin{table}[t]
    \centering
    \caption{Parameters used in the optimization experiments, by optimizer.}
    \begin{tabular}{c|c|c}
    \hline
        Approach & Parameter & Value\\ \hline
        \multirow{5}*{All} & Inference steps & 1\\
        & Guidance scale & 0\\ 
        & Image size & $512\times512$ \\
        & $(a,b)$ & $\{(1,0),(0.5,0.5),(0,1)\}$  \\
        & Time frame & 1000 seconds  \\\hline
        \multirow{2}*{sep-CMA-ES} & Population size & 20\\ 
        & Sigma & 0.5\\ \hline
        \multirow{4}*{Adam} & Learning rate & $5\times10^{-3}$\\
        & Epsilon & $10^{-8}$\\ 
        & Weight decay & $10^{-5}$\\ 
        & ($\beta_1$, $\beta_2$) & (0.85, 0.98)\\ \hline
    \end{tabular}
    \label{tab:parameters}
\end{table}

Execution time depends on the hardware and software environment. For transparency, the computational resources used in our experiments are listed in Table~\ref{tab:computational_resources}.

\begin{table}[t]
    \centering
    \caption{Hardware and software specifications used in the SDXL Turbo optimization experiments.}
    \begin{tabular}{c|c}
    \hline
        Component & Specification\\ \hline
        CPU & Intel\textregistered\ Xeon\textregistered\ Silver 4314 @ 2.40GHz\\
        GPU & NVIDIA RTX A6000 48GB\\
        RAM & $8 \times 32$GB @ 3200MHz\\
        Operating System & Ubuntu 22.04.2 LTS\\ \hline
    \end{tabular}
    \label{tab:computational_resources}
\end{table}

The first three parameters in Table~\ref{tab:parameters} are specific to SDXL Turbo. Since the model is designed to produce high-quality images in one to four inference steps, we used a single inference step to leverage fast generation in the optimization loop, which produced thousands of images. The guidance scale and image size were set to their default values of 0 and $512\times512$, respectively. Moreover, because the fitness function balances the two evaluation metrics (Eq.~\ref{eq:fitness}), we defined three experimental settings: (i) aesthetics only, $(a,b)=(1,0)$; (ii) balanced aesthetics and alignment, $(a,b)=(0.5,0.5)$; and (iii) alignment only, $(a,b)=(0,1)$.

For quantitative assessment, we compare LAION Aesthetic Predictor V2, CLIPScore, and the resulting fitness values. The aesthetic score nominally ranges from 1 to 10, although the linear regressor may output values outside this interval, whereas CLIPScore (cosine similarity) ranges from $-1$ to $1$. To keep the fitness value ideally within $[0,1]$, we normalize the aesthetic score and CLIPScore using two manually selected constants based on the maximum values observed in our experiments with EIGO. We used 10 for the aesthetic score and 0.5 for CLIPScore.

\section{Experimental Results}
\label{sec:results}

In this section, we present and discuss the obtained experimental results. Table~\ref{tab:results} reports quantitative outcomes of prompt-embedding optimization on SDXL Turbo, comparing the no-optimization baseline with Adam and sep-CMA-ES under three fitness weightings, $(a,b)\in\{(1,0),(0.5,0.5),(0,1)\}$. For each setting, the table lists the mean, maximum, and standard deviation of the aesthetic score, CLIPScore, and fitness over the evaluation set, along with the percent change relative to the SDXL Turbo baseline under the same $(a,b)$. The table also includes the number of prompts for which each optimizer achieved the highest fitness.

\begin{table}[t]
\centering
\scriptsize
\setlength{\tabcolsep}{2.2pt}
\caption{Results comparison between SDXL Turbo with no optimization (baseline) and the optimized versions using Adam and sep-CMA-ES, compared across weightings \((a,b)\) for LAION Aesthetic V2, CLIPScore, fitness, and number of prompts where the highest fitness score was attained. The columns report the mean, standard deviation, maximum, and percentage change relative to the baseline. The highest mean and percentage change per metric for each experimentation scenario is highlighted in bold, as well as the algorithm with the highest average fitness.}
\begin{tabular}{lccrrrrrrrrrrrrr}
\toprule
\multirow{2}{*}{Algorithm} & \multirow{2}{*}{$a$} & \multirow{2}{*}{$b$} &
\multicolumn{4}{c}{LAION Aesthetic V2 [1,10] $\uparrow$} &
\multicolumn{4}{c}{CLIPScore [-1, 1] $\uparrow$} &
\multicolumn{4}{c}{Fitness [0,1] $\uparrow$} &
\multirow{2}{*}{Wins [0--36] $\uparrow$} \\
\cmidrule(lr){4-7}\cmidrule(lr){8-11}\cmidrule(lr){12-15}
& & & Avg. & Std. & Max & $\Delta$ base (\%) & Avg. & Std. & Max & $\Delta$ base (\%) & Avg. & Std. & Max & $\Delta$ base (\%) & \# prompts \\
\midrule
SDXL Turbo (no optimization) & 1   & 0   & 5.75 & 0.58 & 6.76 & 0.00  & \textbf{0.2778} & 0.0508 & 0.4087 & \textbf{0.00}  & 0.5751 & 0.0581 & 0.6762 & 0.00  & 0 \\
Adam                          & 1   & 0   & 7.12 & 0.73 & 8.19 & 23.83 & 0.2499 & 0.0542 & 0.3976 & -10.03 & 0.7121 & 0.0734 & 0.8189 & 23.83 & 0 \\
\textbf{sep-CMA-ES}                    & 1   & 0   & \textbf{8.32} & 0.52 & 9.16 & \textbf{44.72} & 0.2141 & 0.0583 & 0.3586 & -22.91 & \textbf{0.8323} & 0.0524 & 0.9160 & \textbf{44.72} & \textbf{36} \\
\midrule
SDXL Turbo (no optimization) & 0.5 & 0.5 & 5.75 & 0.58 & 6.76 & 0.00  & 0.2778 & 0.0508 & 0.4087 & 0.00  & 0.5653 & 0.0621 & 0.7426 & 0.00  & 0 \\
Adam                          & 0.5 & 0.5 & 6.16 & 0.62 & 7.45 & 7.18  & 0.3159 & 0.0577 & 0.4582 & 13.72 & 0.6241 & 0.0664 & 0.7898 & 10.39 & 1 \\
\textbf{sep-CMA-ES }                   & 0.5 & 0.5 & \textbf{7.27} & 0.63 & 8.48 & \textbf{26.44} & \textbf{0.3696} & 0.0764 & 0.5112 & \textbf{33.07} & \textbf{0.7332} & 0.0668 & 0.8855 & \textbf{29.70} & \textbf{35} \\
\midrule
SDXL Turbo (no optimization) & 0   & 1   & \textbf{5.75} & 0.58 & 6.76 & \textbf{0.00}  & 0.2778 & 0.0508 & 0.4087 & 0.00  & 0.5556 & 0.1016 & 0.8173 & 0.00  & 0 \\
Adam                          & 0   & 1   & 5.70 & 0.58 & 6.60 & -0.95 & 0.3517 & 0.0756 & 0.5385 & 26.62 & 0.7035 & 0.1512 & 1.0770 & 26.62 & 4 \\
\textbf{sep-CMA-ES}                    & 0   & 1   & 5.62 & 0.61 & 6.89 & -2.25 & \textbf{0.3977} & 0.0680 & 0.5439 & \textbf{43.17} & \textbf{0.7954} & 0.1361 & 1.0879 & \textbf{43.17} & \textbf{32} \\
\bottomrule
\end{tabular}
\label{tab:results}
\end{table}


sep-CMA-ES achieved a higher mean fitness across all weight settings. In the aesthetics-only setting, sep-CMA-ES attained a mean fitness of 0.8323, corresponding to a 44.72\% improvement over the baseline (0.5751), whereas Adam achieved a 23.83\% improvement with a mean fitness of 0.7121. With equal weights on aesthetics and alignment, sep-CMA-ES improved fitness by 29.70\% (0.7332), driven by a 26.44\% increase in the aesthetic score and a 33.07\% increase in CLIPScore. In the same setting, Adam yielded a 10.39\% fitness improvement (0.6241). Finally, in the alignment-only setting, sep-CMA-ES again showed a clear advantage, achieving a 43.16\% increase in fitness (0.7954), compared to Adam's 26.62\% (0.7035) improvement.

\begin{figure}[t]
  \centering
  \includegraphics[width=14cm]{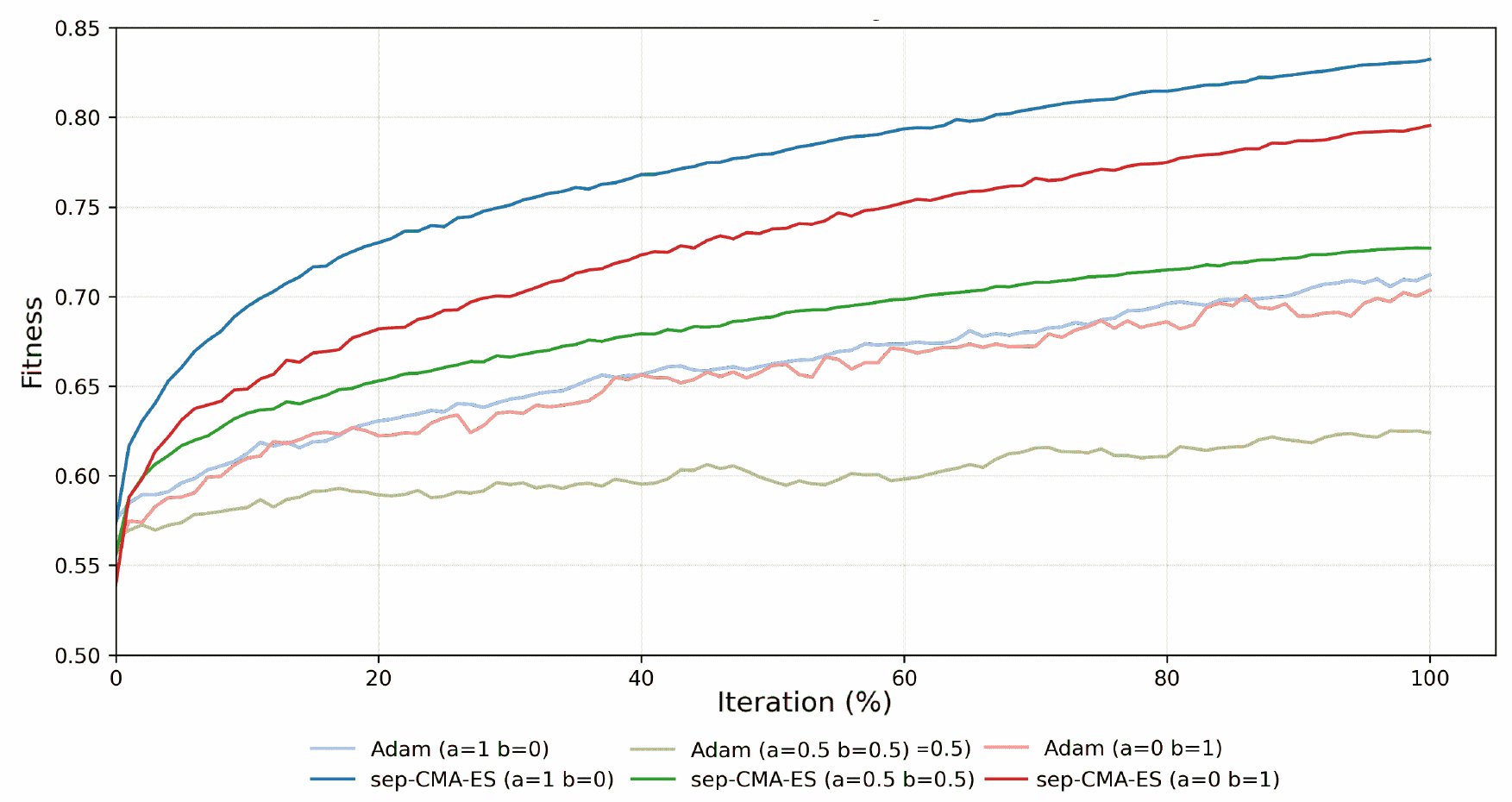}
  \caption{Mean fitness evolution comparison between Adam and sep-CMA-ES across all of the $(a,b)$ combinations.}
  \label{fig:fitness_evol}
\end{figure}

Figure~\ref{fig:fitness_evol} shows the mean fitness over the course of optimization for all weight settings.

Both curves remain on an upward trend, suggesting that a larger iteration budget could yield higher fitness values, particularly for sep-CMA-ES. Nevertheless, the plots also show a consistent advantage for sep-CMA-ES in all settings.

\begin{figure}[htbp]
  \centering
  \includegraphics[width=8.5cm]{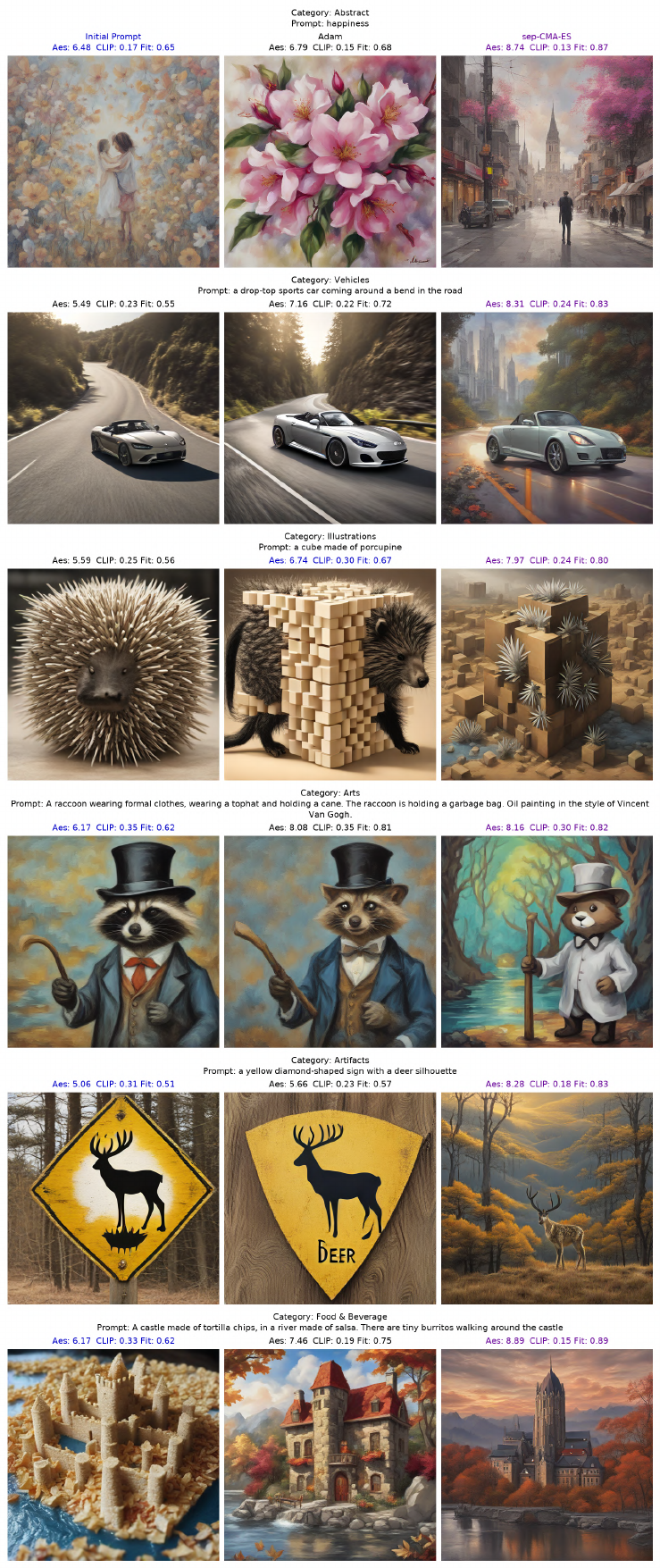}
  \caption{Final outputs from baseline SDXL Turbo, Adam, and sep-CMA-ES for 6 prompts in the aesthetics-only setting. Rows correspond to prompts and columns to methods, with aesthetic, CLIP, and fitness scores above each image; purple marks the highest-fitness image, while red or blue mark the best aesthetic or CLIPScore when they do not match the fitness optimum.}
  \label{fig:sample_results_grid_a100_b0}
\end{figure}

\begin{figure}[htbp]
  \centering
  \includegraphics[width=8.5cm]{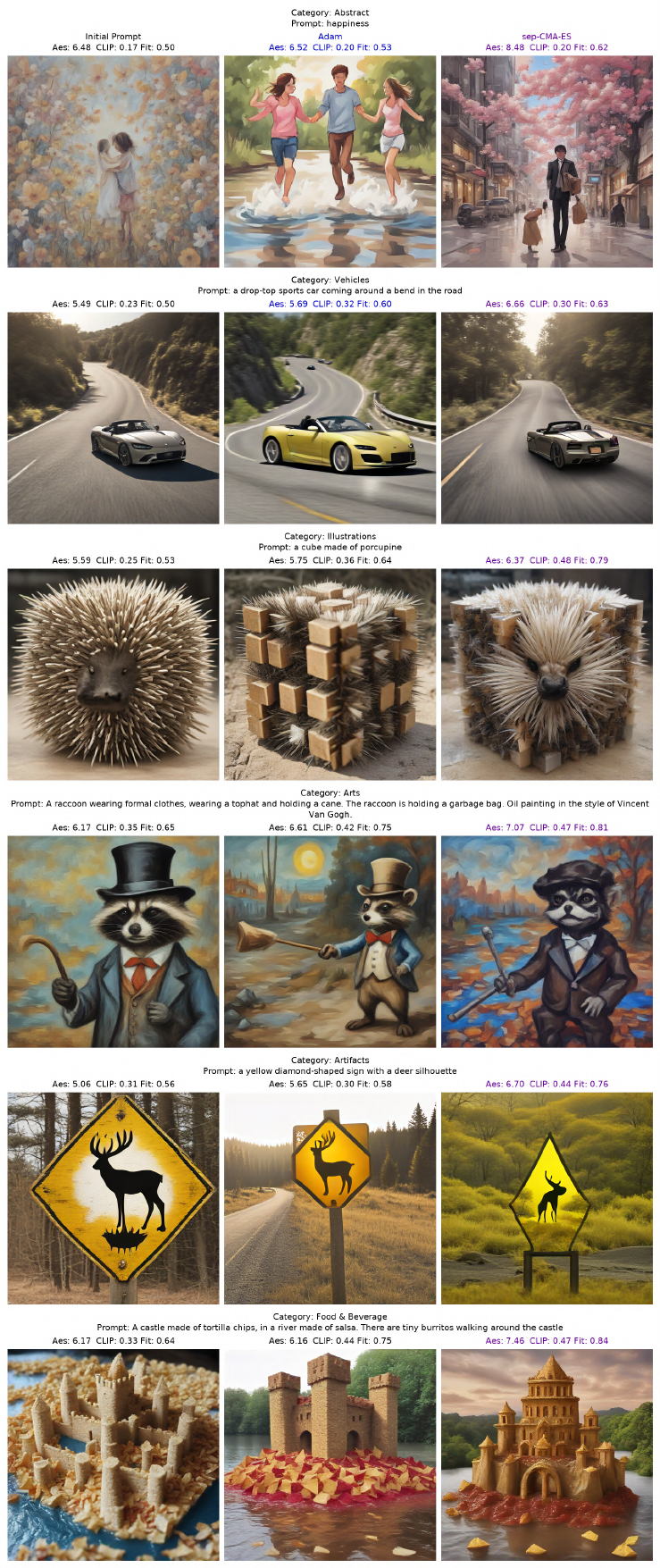}
  \caption{Final outputs from baseline SDXL Turbo, Adam, and sep-CMA-ES for 6 prompts in the balanced setting. Rows correspond to prompts and columns to methods, with aesthetic, CLIP, and fitness scores above each image; purple marks the highest-fitness image, while red or blue mark the best aesthetic or CLIPScore when they do not match the fitness optimum.}
  \label{fig:sample_results_grid_a50_b50}
\end{figure}

\begin{figure}[htbp]
  \centering
  \includegraphics[width=8.5cm]{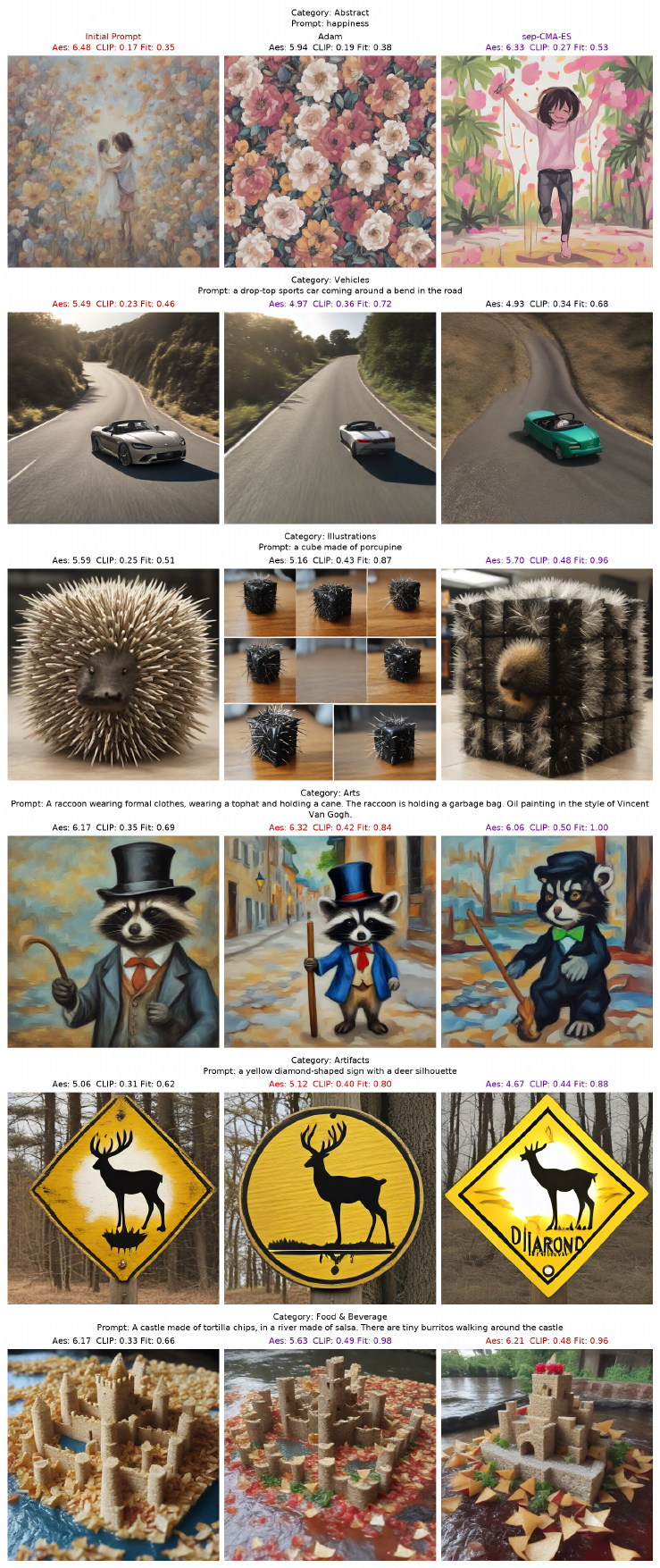}
  \caption{Final outputs from baseline SDXL Turbo, Adam, and sep-CMA-ES for 6 prompts in the alignment-only setting. Rows correspond to prompts and columns to methods, with aesthetic, CLIP, and fitness scores above each image; purple marks the highest-fitness image, while red or blue mark the best aesthetic or CLIPScore when they do not match the fitness optimum.}
  \label{fig:sample_results_grid_a0_b100}
\end{figure}

A visual comparison of the final outputs for 6 example prompts is shown in Figures~\ref{fig:sample_results_grid_a100_b0}--\ref{fig:sample_results_grid_a0_b100}. Each figure contains three columns (one per method), with rows corresponding to prompts. The aesthetic, CLIPScore, and fitness values are indicated above each image. Values are highlighted in purple when the image achieves the highest fitness for that prompt; when the fitness winner differs from the best aesthetic or CLIPScore, the values are highlighted in red for best aesthetics and blue for best CLIPScore.

In the presented examples, the baseline images are often less detailed and use simpler lighting. In the aesthetics-only setting, Adam tends to remain closer to the baseline, whereas sep-CMA-ES explores more diverse solutions, often introducing different scenarios with additional details. This divergence from the baseline is expected in this setting since prompt alignment is not included in the objective. In the balanced setting, both methods produced outputs closer to the baseline. In the alignment-only setting, by not considering aesthetics, both optimizers yield solutions with more literal representations of the prompt but also with more visual artifacts. 

\begin{figure}[htbp]
  \centering
  \includegraphics[width=12cm]{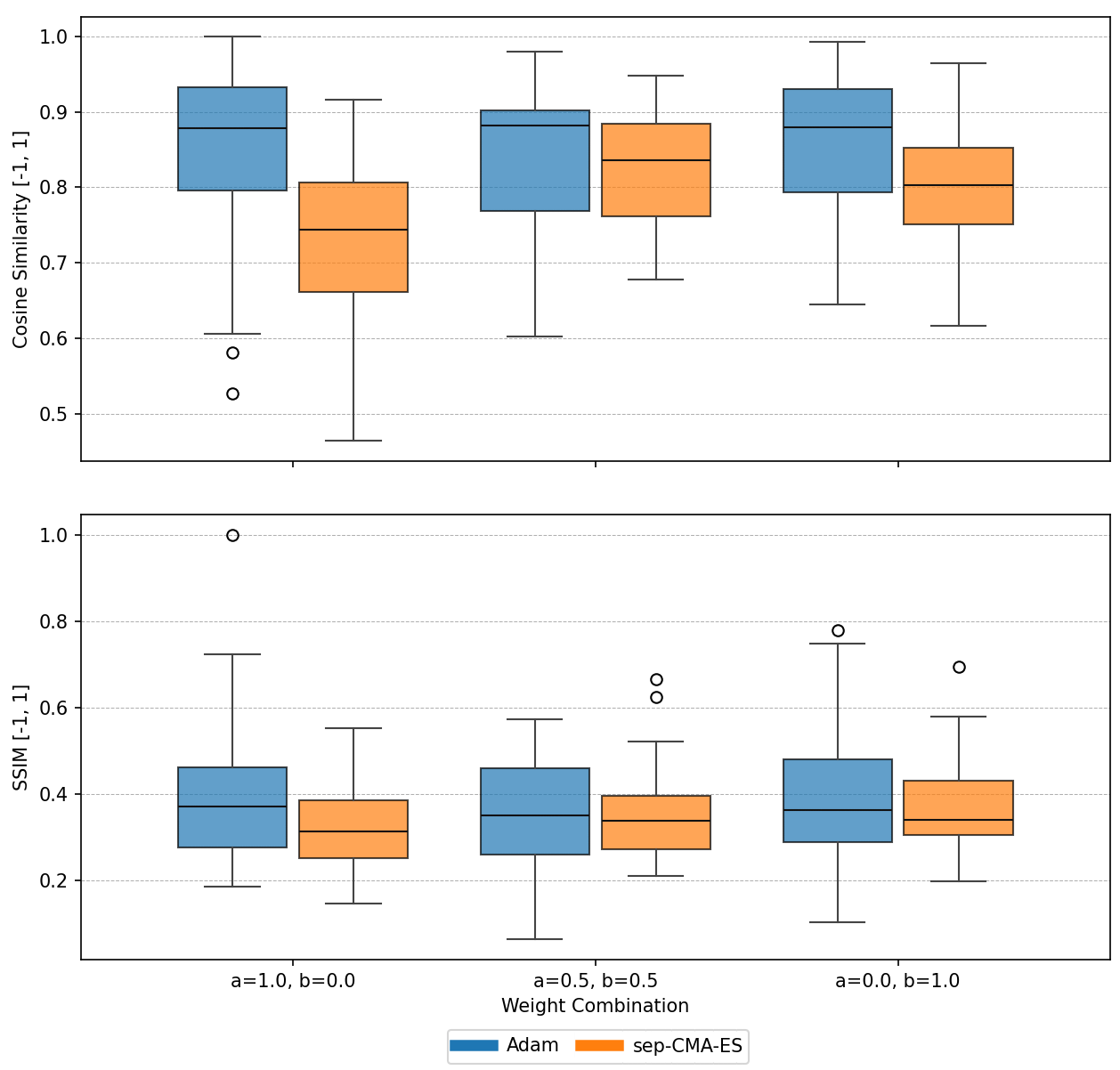}
  \caption{Box plot of cosine similarity (top) and SSIM (bottom) between the final image for each method and the no-optimization baseline for the 36 prompts, grouped by weight setting: (i) aesthetics only, (ii) balanced, and (iii) alignment only.}
  \label{fig:cosine_ssim}
\end{figure}

To calculate the similarity to the baseline we compute cosine similarity and the Structural Similarity Index Measure (SSIM) between the final image produced by each optimizer and the corresponding baseline image. Figure~\ref{fig:cosine_ssim} shows aggregated results for both metrics grouped by weight setting.

Across all settings, sep-CMA-ES shows, on average, lower similarity to the baseline under both measures. As expected, the lowest similarity scores occur in the aesthetics-only setting, where the optimizers are free to deviate more strongly from the baseline.

Overall, sep-CMA-ES outperformed Adam across the prompt-embedding search optimization experiments according to the defined evaluation metrics. sep-CMA-ES consistently achieved higher fitness values and, depending on the weight setting, improved either aesthetics, alignment, or both. It also explored farther from the baseline starting point, which is consistent with the similarity analysis. In terms of computational resources, Adam required 39.3\,GB of VRAM on our system (Table~\ref{tab:computational_resources}), whereas sep-CMA-ES required 17.6\,GB, that is, less than half. This difference is largely explained by the memory cost of backpropagation and gradient tracking in Adam.

In summary, these results support evolutionary optimization as an effective and cost-efficient approach for inference time control in diffusion-based image generation. At the same time, further work is required to better understand the behavior of these optimizers beyond prompt-embedding search and under alternative objectives and evaluators.

However, this approach has limitations, particularly in terms of runtime. On average, sep-CMA-ES required 15\,min to complete 100 generations with a population of 20, which is substantially slower than the $\sim$0.3\,s required to generate a single image without optimization. This overhead is inherent to the iterative loop that repeatedly generates an image, evaluates it, and updates the parameters. Improving algorithmic efficiency and parallelization could therefore enhance practicality, for example by decoupling image generation and evaluation in evolutionary runs and using multiple instances of the generator and evaluators to speed up each generation.

Another challenge is that optimization is sensitive to hyperparameters (e.g., population, mutation/step size, and learning rate). An in-depth parameter study would allow a systematic understanding of their influence on the convergence behavior, stability, and solution quality, enabling the identification of optimal configurations for different objective weightings and generative model settings. Such an investigation could also focus on parameter auto-tuning, increasing the usability and interoperability between generative models and optimization methods.

\section{Conclusion and Future Work}
\label{sec:conclusions}

This work presents a comparison of sep-CMA-ES and Adam for embedding-space exploration via inference-time prompt-embedding optimization in diffusion-based image generation. We optimize prompt embedding vectors to improve both image aesthetics and prompt--image alignment, using a weighted objective that combines the LAION Aesthetic Predictor V2 and CLIPScore. Experiments with Stable Diffusion XL Turbo show that sep-CMA-ES consistently achieves higher objective values across all weight settings while using less than half of the VRAM required by Adam. These results support evolutionary optimizers as an effective approach for embedding-space search, enabling controllable improvements without retraining or architectural changes. We release the EIGO engine to facilitate replication and further experimentation.

Future work should examine a broader set of optimizers. We selected sep-CMA-ES as a simple and scalable CMA-ES variant, but alternatives such as LM-CMA-ES could capture cross-coordinate dependencies more effectively while remaining cheaper than full CMA-ES~\cite{losh2014}. Other evolutionary methods, including Particle Swarm Optimization (PSO) and hybrid evolutionary/gradient-based approaches, may offer different trade-offs between exploration and computational cost~\cite{kennedy1995,Correia2024}. Extending the study to additional generators, such as FLUX~\cite{labs2025flux1kontextflowmatching}, PixArt~\cite{chen2023}, and QwenImage~\cite{wu2025}, would further clarify how well inference-time embedding optimization generalizes across model families.

Another promising direction is human-in-the-loop evaluation~\cite{cristiano2017}, which could improve optimization for complex or abstract prompts that are difficult to assess reliably with CLIPScore alone and may be vulnerable to reward exploitation. In parallel, we will continue evolving EIGO as a modular framework in which users can mix and match generators, evaluators, and optimizers to support inference-time optimization across a wider range of image-generative models and objectives.

\paragraph{Disclaimer.} Large language models were used for language editing (grammar, style, and clarity). All technical and scientific content, including claims, experimental design, results, and conclusions, is the responsibility of the authors.

\section*{Acknowledgments}

This work is funded by national funds through FCT – Foundation for Science and Technology, I.P., within the scope of the research unit UID/00326 - Centre for Informatics and Systems of the University of Coimbra, and through the Portuguese Recovery and Resilience Plan (PRR) through project C645008882-00000055, Center for Responsible AI.

\bibliographystyle{unsrt}  
\bibliography{references}  

@article{Rombach2022,
archivePrefix = {arXiv},
arxivId = {2112.10752},
author = {Rombach, Robin and Blattmann, Andreas and Lorenz, Dominik and Esser, Patrick and Ommer, Bjorn},
doi = {10.1109/CVPR52688.2022.01042},
eprint = {2112.10752},
isbn = {9781665469463},
issn = {10636919},
journal = {Proceedings of the IEEE Computer Society Conference on Computer Vision and Pattern Recognition},
pages = {10674--10685},
title = {{High-Resolution Image Synthesis with Latent Diffusion Models}},
volume = {2022-June},
year = {2022}
}

@article{Li2025review,
author = {Li, Jun and Zhang, Chenyang and Zhu, Wei and Ren, Yawei},
doi = {10.1007/s40745-024-00544-1},
issn = {21985812},
journal = {Annals of Data Science},
number = {1},
pages = {141--170},
publisher = {Springer Berlin Heidelberg},
title = {{A Comprehensive Survey of Image Generation Models Based on Deep Learning}},
url = {https://doi.org/10.1007/s40745-024-00544-1},
volume = {12},
year = {2025}
}

@misc{Schuhmann2022,
   author = {Christoph Schuhmann},
   journal = {LAION},
   month = {8},
   title = {LAION-Aesthetics},
   url = {https://laion.ai/blog/laion-aesthetics/},
   year = {2022}
}

@inproceedings{xu2023,
author = {Xu, Jiazheng and Liu, Xiao and Wu, Yuchen and Tong, Yuxuan and Li, Qinkai and Ding, Ming and Tang, Jie and Dong, Yuxiao},
title = {ImageReward: learning and evaluating human preferences for text-to-image generation},
year = {2023},
publisher = {Curran Associates Inc.},
address = {Red Hook, NY, USA},
booktitle = {Proceedings of the 37th International Conference on Neural Information Processing Systems},
articleno = {700},
numpages = {33},
location = {New Orleans, LA, USA},
series = {NIPS '23}
}

@article{hessel2021,
  title={CLIPScore: A Reference-free Evaluation Metric for Image Captioning},
  author={Jack Hessel and Ari Holtzman and Maxwell Forbes and Ronan Le Bras and Yejin Choi},
  journal={ArXiv},
  year={2021},
  volume={abs/2104.08718},
  url={https://api.semanticscholar.org/CorpusID:233296711}
}

@article{mirza2014,
      title={Conditional Generative Adversarial Nets}, 
      author={Mehdi Mirza and Simon Osindero},
      year={2014},
      journal={ArXiv},
      volume={abs/1411.1784},
      eprint={1411.1784},
      archivePrefix={arXiv},
      primaryClass={cs.LG},
      url={https://arxiv.org/abs/1411.1784}, 
}

@INPROCEEDINGS{park2019,
  author={Park, Taesung and Liu, Ming-Yu and Wang, Ting-Chun and Zhu, Jun-Yan},
  booktitle={2019 IEEE/CVF Conference on Computer Vision and Pattern Recognition (CVPR)}, 
  title={Semantic Image Synthesis With Spatially-Adaptive Normalization}, 
  year={2019},
  volume={},
  number={},
  pages={2332-2341},
  doi={10.1109/CVPR.2019.00244}}

@INPROCEEDINGS{peebles2023,
  author={Peebles, William and Xie, Saining},
  booktitle={2023 IEEE/CVF International Conference on Computer Vision (ICCV)}, 
  title={Scalable Diffusion Models with Transformers}, 
  year={2023},
  volume={},
  number={},
  pages={4172-4182},
  doi={10.1109/ICCV51070.2023.00387}}

@article{imagenteamgoogle2024imagen3,
  title={Imagen 3},
  author={Imagen-Team-Google et al.},
  year={2024},
  journal={arXiv preprint arXiv:2408.07009},
  url={https://arxiv.org/abs/2408.07009}
}

@inproceedings{esser2024,
author = {Esser, Patrick and Kulal, Sumith and Blattmann, Andreas and Entezari, Rahim and M\"{u}ller, Jonas and Saini, Harry and Levi, Yam and Lorenz, Dominik and Sauer, Axel and Boesel, Frederic and Podell, Dustin and Dockhorn, Tim and English, Zion and Rombach, Robin},
title = {Scaling rectified flow transformers for high-resolution image synthesis},
year = {2024},
publisher = {JMLR.org},
booktitle = {Proceedings of the 41st International Conference on Machine Learning},
articleno = {503},
numpages = {28},
location = {Vienna, Austria},
series = {ICML'24}
}

@article{labs2025flux1kontextflowmatching,
      title={FLUX.1 Kontext: Flow Matching for In-Context Image Generation and Editing in Latent Space}, 
      author={Black Forest Labs and Stephen Batifol and Andreas Blattmann and Frederic Boesel and Saksham Consul and Cyril Diagne and Tim Dockhorn and Jack English and Zion English and Patrick Esser and Sumith Kulal and Kyle Lacey and Yam Levi and Cheng Li and Dominik Lorenz and Jonas Müller and Dustin Podell and Robin Rombach and Harry Saini and Axel Sauer and Luke Smith},
      year={2025},
      journal={ArXiv},
      volume={abs/2506.15742},
      eprint={2506.15742},
      archivePrefix={arXiv},
      primaryClass={cs.GR},
      url={https://arxiv.org/abs/2506.15742}, 
}

@inbook{sobania2025,
author = {Sobania, Dominik and Briesch, Martin and Rothlauf, Franz},
title = {ImageBreeder: Guiding Diffusion Models with Evolutionary Computation},
year = {2025},
isbn = {9798400714658},
publisher = {Association for Computing Machinery},
address = {New York, NY, USA},
url = {https://doi.org/10.1145/3712256.3726439},
booktitle = {Proceedings of the Genetic and Evolutionary Computation Conference},
pages = {463–471},
numpages = {9}
}

@inproceedings{salvenmoser2025,
author = {Salvenmoser, Marcel and Affenzeller, Michael},
title = {Evolving the Embedding Space of Diffusion Models in the Field of Visual Arts},
year = {2025},
isbn = {978-3-031-90166-9},
publisher = {Springer-Verlag},
address = {Berlin, Heidelberg},
url = {https://doi.org/10.1007/978-3-031-90167-6_27},
doi = {10.1007/978-3-031-90167-6_27},
booktitle = {Artificial Intelligence in Music, Sound, Art and Design: 14th International Conference, EvoMUSART 2025, Held as Part of EvoStar 2025, Trieste, Italy, April 23–25, 2025, Proceedings},
pages = {402–416},
numpages = {15},
location = {Trieste, Italy}
}

@inproceedings{martins2023metaprompter,
	address = {Cham},
	author = {Martins, Tiago and Cunha, Jo{\~ a}o M. and Correia, Jo{\~ a}o and Machado, Penousal},
	booktitle = {Artificial {Intelligence} in {Music}, {Sound}, {Art} and {Design}},
	editor = {Johnson, Colin and Rodriguez-Fernandez, Nereida and Rebelo, Sergio M.},
	year = {2023},
	pages = {180--195},
	organization = {Springer Nature Switzerland},
	title = {Towards the {Evolution} of {Prompts} with {MetaPrompter}},
}

@Inbook{Correia2024,
author="Correia, Jo{\~a}o
and Baeta, Francisco
and Martins, Tiago",
title="Evolutionary Generative Models",
bookTitle="Handbook of Evolutionary Machine Learning",
year="2024",
publisher="Springer Nature Singapore",
address="Singapore",
pages="283--329",
isbn="978-981-99-3814-8",
doi="10.1007/978-981-99-3814-8_10",
url="https://doi.org/10.1007/978-981-99-3814-8_10"
}

@INPROCEEDINGS{zhang2024,
  author={Zhang, Sixian and Wang, Bohan and Wu, Junqiang and Li, Yan and Gao, Tingting and Zhang, Di and Wang, Zhongyuan},
  booktitle={2024 IEEE/CVF Conference on Computer Vision and Pattern Recognition (CVPR)}, 
  title={Learning Multi-Dimensional Human Preference for Text-to-Image Generation}, 
  year={2024},
  volume={},
  number={},
  pages={8018-8027},
  doi={10.1109/CVPR52733.2024.00766}}

@INPROCEEDINGS{liang2024,
  author={Liang, Youwei and He, Junfeng and Li, Gang and Li, Peizhao and Klimovskiy, Arseniy and Carolan, Nicholas and Sun, Jiao and Pont-Tuset, Jordi and Young, Sarah and Yang, Feng and Ke, Junjie and Dvijotham, Krishnamurthy Dj and Collins, Katherine M. and Luo, Yiwen and Li, Yang and Kohlhoff, Kai J and Ramachandran, Deepak and Navalpakkam, Vidhya},
  booktitle={2024 IEEE/CVF Conference on Computer Vision and Pattern Recognition (CVPR)}, 
  title={Rich Human Feedback for Text-to-Image Generation}, 
  year={2024},
  volume={},
  number={},
  pages={19401-19411},
  doi={10.1109/CVPR52733.2024.01835}}

@InProceedings{hu2023,
    author    = {Hu, Yushi and Liu, Benlin and Kasai, Jungo and Wang, Yizhong and Ostendorf, Mari and Krishna, Ranjay and Smith, Noah A.},
    title     = {TIFA: Accurate and Interpretable Text-to-Image Faithfulness Evaluation with Question Answering},
    booktitle = {Proceedings of the IEEE/CVF International Conference on Computer Vision (ICCV)},
    month     = {October},
    year      = {2023},
    pages     = {20406-20417}
}

@inproceedings{yuval2023,
author = {Kirstain, Yuval and Polyak, Adam and Singer, Uriel and Matiana, Shahbuland and Penna, Joe and Levy, Omer},
title = {Pick-a-Pic: an open dataset of user preferences for text-to-image generation},
year = {2023},
publisher = {Curran Associates Inc.},
address = {Red Hook, NY, USA},
booktitle = {Proceedings of the 37th International Conference on Neural Information Processing Systems},
articleno = {1594},
numpages = {12},
location = {New Orleans, LA, USA},
series = {NIPS '23}
}

@inproceedings{costa2023,
author = {Costa, Victor and Louren\c{c}o, Nuno and Correia, Jo\~{a}o and Machado, Penousal},
title = {Exploring Generative Adversarial Networks for Text-to-Image Generation with Evolution Strategies},
year = {2023},
isbn = {9798400701207},
publisher = {Association for Computing Machinery},
address = {New York, NY, USA},
url = {https://doi.org/10.1145/3583133.3590549},
doi = {10.1145/3583133.3590549},
booktitle = {Proceedings of the Companion Conference on Genetic and Evolutionary Computation},
pages = {271–274},
numpages = {4},
location = {Lisbon, Portugal},
series = {GECCO '23 Companion}
}

@INPROCEEDINGS{haruka2023,
  author={Kobayashi, Haruka and Pindur, Adam Kotaro and Venkatesh, Suryanarayanan Nagar Anthel and Iba, Hitoshi},
  booktitle={2023 IEEE International Conference on Systems, Man, and Cybernetics (SMC)}, 
  title={Image Generation with Diffusion Model by Interactive Evolutionary Computation}, 
  year={2023},
  volume={},
  number={},
  pages={2984-2990},
  doi={10.1109/SMC53992.2023.10394041}}

@inproceedings{clare2023,
author = {Clare, Luana and Correia, Jo\~{a}o},
title = {Generating Adversarial Examples through Latent Space Exploration of Generative Adversarial Networks},
year = {2023},
isbn = {9798400701207},
publisher = {Association for Computing Machinery},
address = {New York, NY, USA},
url = {https://doi.org/10.1145/3583133.3596392},
doi = {10.1145/3583133.3596392},
booktitle = {Proceedings of the Companion Conference on Genetic and Evolutionary Computation},
pages = {1760–1767},
numpages = {8},
location = {Lisbon, Portugal},
series = {GECCO '23 Companion}
}

@INPROCEEDINGS{tran2023,
  author={Tran, Khoi Dinh and Bui, Dat Viet and Luong, Ngoc Hoang},
  booktitle={2023 International Conference on Multimedia Analysis and Pattern Recognition (MAPR)}, 
  title={Evolving Prompts for Synthetic Image Generation with Genetic Algorithm}, 
  year={2023},
  volume={},
  number={},
  pages={1-6},
  doi={10.1109/MAPR59823.2023.10288925}}

@inproceedings{hao2023,
 author = {Hao, Yaru and Chi, Zewen and Dong, Li and Wei, Furu},
 booktitle = {Advances in Neural Information Processing Systems},
 editor = {A. Oh and T. Naumann and A. Globerson and K. Saenko and M. Hardt and S. Levine},
 pages = {66923--66939},
 publisher = {Curran Associates, Inc.},
 title = {Optimizing Prompts for Text-to-Image Generation},
 url = {https://proceedings.neurips.cc/paper_files/paper/2023/file/d346d91999074dd8d6073d4c3b13733b-Paper-Conference.pdf},
 volume = {36},
 year = {2023}
}

@inproceedings{li2025,
author = {Li, WeiJie and Wang, Jin and Zhang, Xuejie},
title = {PROMPTIST: Automated Prompt Optimization for Text-to-Image Synthesis},
year = {2024},
isbn = {978-981-97-9433-1},
publisher = {Springer-Verlag},
address = {Berlin, Heidelberg},
url = {https://doi.org/10.1007/978-981-97-9434-8_23},
doi = {10.1007/978-981-97-9434-8_23},
booktitle = {Natural Language Processing and Chinese Computing: 13th National CCF Conference, NLPCC 2024, Hangzhou, China, November 1–3, 2024, Proceedings, Part II},
pages = {295–306},
numpages = {12},
location = {Hangzhou, China}
}

@INPROCEEDINGS{wong2023,
  author={Wong, Melvin and Ong, Yew-Soon and Gupta, Abhishek and Bali, Kavitesh Kumar and Chen, Caishun},
  booktitle={2023 IEEE Conference on Artificial Intelligence (CAI)}, 
  title={Prompt Evolution for Generative AI: A Classifier-Guided Approach}, 
  year={2023},
  volume={},
  number={},
  pages={226-229},
  doi={10.1109/CAI54212.2023.00105}}

@inproceedings{wang2024,
author = {Wang, Zhijie and Huang, Yuheng and Song, Da and Ma, Lei and Zhang, Tianyi},
title = {PromptCharm: Text-to-Image Generation through Multi-modal Prompting and Refinement},
year = {2024},
isbn = {9798400703300},
publisher = {Association for Computing Machinery},
address = {New York, NY, USA},
url = {https://doi.org/10.1145/3613904.3642803},
doi = {10.1145/3613904.3642803},
booktitle = {Proceedings of the 2024 CHI Conference on Human Factors in Computing Systems},
articleno = {185},
numpages = {21},
location = {Honolulu, HI, USA},
series = {CHI '24}
}

@article{Wu2023,
archivePrefix = {arXiv},
arxivId = {2212.08698},
author = {Wu, Qiucheng and Liu, Yujian and Zhao, Handong and Kale, Ajinkya and Bui, Trung and Yu, Tong and Lin, Zhe and Zhang, Yang and Chang, Shiyu},
doi = {10.1109/CVPR52729.2023.00189},
eprint = {2212.08698},
isbn = {9798350301298},
issn = {10636919},
journal = {Proceedings of the IEEE Computer Society Conference on Computer Vision and Pattern Recognition},
pages = {1900--1910},
title = {{Uncovering the Disentanglement Capability in Text-to-Image Diffusion Models}},
volume = {2023-June},
year = {2023}
}

@article{Yu2024,
archivePrefix = {arXiv},
arxivId = {2404.01154},
author = {Yu, Hu and Luo, Hao and Wang, Fan and Zhao, Feng},
eprint = {2404.01154},
title = {{Uncovering the Text Embedding in Text-to-Image Diffusion Models}},
url = {http://arxiv.org/abs/2404.01154},
year = {2024},
journal={ArXiv},
volume={abs/2404.01154},
}

@article{hansen2001,
  title={Completely derandomized self-adaptation in evolution strategies},
  author={Hansen, Nikolaus and Ostermeier, Andreas},
  journal={Evolutionary Computation},
  volume={9},
  number={2},
  pages={159--195},
  year={2001},
  publisher={MIT Press},
  doi={10.1162/106365601750190398}
}

@InProceedings{ros2008,
author="Ros, Raymond
and Hansen, Nikolaus",
editor="Rudolph, G{\"u}nter
and Jansen, Thomas
and Beume, Nicola
and Lucas, Simon
and Poloni, Carlo",
title="A Simple Modification in CMA-ES Achieving Linear Time and Space Complexity",
booktitle="Parallel Problem Solving from Nature -- PPSN X",
year="2008",
publisher="Springer Berlin Heidelberg",
address="Berlin, Heidelberg",
pages="296--305",
isbn="978-3-540-87700-4"
}

@article{Kingma2014,
  title={Adam: A Method for Stochastic Optimization},
  author={Diederik P. Kingma and Jimmy Ba},
  journal={CoRR},
  year={2014},
  volume={abs/1412.6980},
  url={https://api.semanticscholar.org/CorpusID:6628106}
}

@InProceedings{sauer2025,
author="Sauer, Axel
and Lorenz, Dominik
and Blattmann, Andreas
and Rombach, Robin",
editor="Leonardis, Ale{\v{s}}
and Ricci, Elisa
and Roth, Stefan
and Russakovsky, Olga
and Sattler, Torsten
and Varol, G{\"u}l",
title="Adversarial Diffusion Distillation",
booktitle="Computer Vision -- ECCV 2024",
year="2025",
publisher="Springer Nature Switzerland",
address="Cham",
pages="87--103",
isbn="978-3-031-73016-0"
}

@article{chen2023,
      title={PixArt-$\alpha$: Fast Training of Diffusion Transformer for Photorealistic Text-to-Image Synthesis}, 
      author={Junsong Chen and Jincheng Yu and Chongjian Ge and Lewei Yao and Enze Xie and Yue Wu and Zhongdao Wang and James Kwok and Ping Luo and Huchuan Lu and Zhenguo Li},
      year={2023},
      journal={ArXiv},
      volume={abs/2310.00426},
      eprint={2310.00426},
      archivePrefix={arXiv},
      primaryClass={cs.CV},
      url={https://arxiv.org/abs/2310.00426}, 
}

@misc{deepfloyd2023if,
  title        = {IF by DeepFloyd},
  author       = {{DeepFloyd Team}},
  year         = {2023},
  howpublished = {\url{https://github.com/deep-floyd/IF}},
  note         = {Accessed: 2025-10-08}
}

@article{podell2023,
      title={SDXL: Improving Latent Diffusion Models for High-Resolution Image Synthesis}, 
      author={Dustin Podell and Zion English and Kyle Lacey and Andreas Blattmann and Tim Dockhorn and Jonas Müller and Joe Penna and Robin Rombach},
      year={2023},
      journal={ArXiv},
      volume={abs/2307.01952},
      eprint={2307.01952},
      archivePrefix={arXiv},
      primaryClass={cs.CV},
      url={https://arxiv.org/abs/2307.01952}, 
}

@inproceedings{losh2014,
author = {Loshchilov, Ilya},
title = {A computationally efficient limited memory CMA-ES for large scale optimization},
year = {2014},
isbn = {9781450326629},
publisher = {Association for Computing Machinery},
address = {New York, NY, USA},
url = {https://doi.org/10.1145/2576768.2598294},
doi = {10.1145/2576768.2598294},
booktitle = {Proceedings of the 2014 Annual Conference on Genetic and Evolutionary Computation},
pages = {397–404},
numpages = {8},
location = {Vancouver, BC, Canada},
series = {GECCO '14}
}

@INPROCEEDINGS{kennedy1995,
  author={Kennedy, J. and Eberhart, R.},
  booktitle={Proceedings of ICNN'95 - International Conference on Neural Networks}, 
  title={Particle swarm optimization}, 
  year={1995},
  volume={4},
  number={},
  pages={1942-1948 vol.4},
  doi={10.1109/ICNN.1995.488968}}

@article{wu2025,
      title={Qwen-Image Technical Report}, 
      author={Chenfei Wu and Jiahao Li and Jingren Zhou and Junyang Lin and Kaiyuan Gao and Kun Yan and Sheng-ming Yin and Shuai Bai and Xiao Xu and Yilei Chen and Yuxiang Chen and Zecheng Tang and Zekai Zhang and Zhengyi Wang and An Yang and Bowen Yu and Chen Cheng and Dayiheng Liu and Deqing Li and Hang Zhang and Hao Meng and Hu Wei and Jingyuan Ni and Kai Chen and Kuan Cao and Liang Peng and Lin Qu and Minggang Wu and Peng Wang and Shuting Yu and Tingkun Wen and Wensen Feng and Xiaoxiao Xu and Yi Wang and Yichang Zhang and Yongqiang Zhu and Yujia Wu and Yuxuan Cai and Zenan Liu},
      year={2025},
      journal={ArXiv},
      volume={abs/2508.02324},
      eprint={2508.02324},
      archivePrefix={arXiv},
      primaryClass={cs.CV},
      url={https://arxiv.org/abs/2508.02324}, 
}

@inproceedings{cristiano2017,
author = {Christiano, Paul F. and Leike, Jan and Brown, Tom B. and Martic, Miljan and Legg, Shane and Amodei, Dario},
title = {Deep reinforcement learning from human preferences},
year = {2017},
isbn = {9781510860964},
publisher = {Curran Associates Inc.},
address = {Red Hook, NY, USA},
booktitle = {Proceedings of the 31st International Conference on Neural Information Processing Systems},
pages = {4302–4310},
numpages = {9},
location = {Long Beach, California, USA},
series = {NIPS'17}
}

@inproceedings{
loshchilov2019,
title={Decoupled Weight Decay Regularization},
author={Ilya Loshchilov and Frank Hutter},
booktitle={International Conference on Learning Representations},
year={2019},
url={https://openreview.net/forum?id=Bkg6RiCqY7},
}

@inproceedings{ruiz2022dreambooth,
  author       = {Nataniel Ruiz and
                  Yuanzhen Li and
                  Varun Jampani and
                  Yael Pritch and
                  Michael Rubinstein and
                  Kfir Aberman},
  title        = {DreamBooth: Fine Tuning Text-to-Image Diffusion Models for Subject-Driven
                  Generation},
  booktitle    = {{IEEE/CVF} Conference on Computer Vision and Pattern Recognition,
                  {CVPR} 2023, Vancouver, BC, Canada, June 17-24, 2023},
  pages        = {22500--22510},
  publisher    = {{IEEE}},
  year         = {2023},
  url          = {https://doi.org/10.1109/CVPR52729.2023.02155},
  doi          = {10.1109/CVPR52729.2023.02155},
  bibsource    = {dblp computer science bibliography, https://dblp.org}
}

@article{ho2022classifierfree,
  author       = {Jonathan Ho and Tim Salimans},
  title        = {Classifier-Free Diffusion Guidance},
  journal      = {CoRR},
  volume       = {abs/2207.12598},
  year         = {2022},
  url          = {https://dblp.org/rec/journals/corr/abs-2207-12598},
  eprinttype   = {arXiv},
  eprint       = {2207.12598},
  archivePrefix= {arXiv},
  primaryClass = {cs.LG},
  bibsource    = {dblp computer science bibliography, https://dblp.org}
}

@inproceedings{meng2021sdedit,
  author       = {Chenlin Meng and
                  Yutong He and
                  Yang Song and
                  Jiaming Song and
                  Jiajun Wu and
                  Jun{-}Yan Zhu and
                  Stefano Ermon},
  title        = {SDEdit: Guided Image Synthesis and Editing with Stochastic Differential
                  Equations},
  booktitle    = {The Tenth International Conference on Learning Representations, {ICLR}
                  2022, Virtual Event, April 25-29, 2022},
  publisher    = {OpenReview.net},
  year         = {2022},
  url          = {https://openreview.net/forum?id=aBsCjcPu\_tE},
  bibsource    = {dblp computer science bibliography, https://dblp.org}
}

@inproceedings{hertz2022prompttoprompt,
  author       = {Amir Hertz and
                  Ron Mokady and
                  Jay Tenenbaum and
                  Kfir Aberman and
                  Yael Pritch and
                  Daniel Cohen{-}Or},
  title        = {Prompt-to-Prompt Image Editing with Cross-Attention Control},
  booktitle    = {The Eleventh International Conference on Learning Representations,
                  {ICLR} 2023, Kigali, Rwanda, May 1-5, 2023},
  publisher    = {OpenReview.net},
  year         = {2023},
  url          = {https://openreview.net/forum?id=\_CDixzkzeyb},
  bibsource    = {dblp computer science bibliography, https://dblp.org}
}

@inproceedings{kim2021diffusionclip,
  author       = {Gwanghyun Kim and
                  Taesung Kwon and
                  Jong Chul Ye},
  title        = {DiffusionCLIP: Text-Guided Diffusion Models for Robust Image Manipulation},
  booktitle    = {{IEEE/CVF} Conference on Computer Vision and Pattern Recognition,
                  {CVPR} 2022, New Orleans, LA, USA, June 18-24, 2022},
  pages        = {2416--2425},
  publisher    = {{IEEE}},
  year         = {2022},
  url          = {https://doi.org/10.1109/CVPR52688.2022.00246},
  doi          = {10.1109/CVPR52688.2022.00246},
  bibsource    = {dblp computer science bibliography, https://dblp.org}
}

@inproceedings{kirstain2023pickapic,
  author       = {Yuval Kirstain and
                  Adam Polyak and
                  Uriel Singer and
                  Shahbuland Matiana and
                  Joe Penna and
                  Omer Levy},
  editor       = {Alice Oh and
                  Tristan Naumann and
                  Amir Globerson and
                  Kate Saenko and
                  Moritz Hardt and
                  Sergey Levine},
  title        = {Pick-a-Pic: An Open Dataset of User Preferences for Text-to-Image
                  Generation},
  booktitle    = {Advances in Neural Information Processing Systems 36: Annual Conference
                  on Neural Information Processing Systems 2023, NeurIPS 2023, New Orleans,
                  LA, USA, December 10 - 16, 2023},
  year         = {2023},
  url          = {http://papers.nips.cc/paper\_files/paper/2023/hash/73aacd8b3b05b4b503d58310b523553c-Abstract-Conference.html},
  bibsource    = {dblp computer science bibliography, https://dblp.org}
}

@article{wu2023hpsv2,
  author       = {Xiaoshi Wu and
                  Yiming Hao and
                  Keqiang Sun and
                  Yixiong Chen and
                  Feng Zhu and
                  Rui Zhao and
                  Hongsheng Li},
  title        = {Human Preference Score v2: {A} Solid Benchmark for Evaluating Human
                  Preferences of Text-to-Image Synthesis},
  journal      = {CoRR},
  volume       = {abs/2306.09341},
  year         = {2023},
  url          = {https://doi.org/10.48550/arXiv.2306.09341},
  doi          = {10.48550/ARXIV.2306.09341},
  eprinttype    = {arXiv},
  eprint       = {2306.09341},
  bibsource    = {dblp computer science bibliography, https://dblp.org}
}

\clearpage
\appendix
\section*{Appendix}
\setcounter{figure}{0}\setcounter{table}{0}
\renewcommand{\thefigure}{A\arabic{figure}}
\renewcommand{\thetable}{A\arabic{table}}

\section{Used Prompts}

\begin{table}[hbt]
\caption{Selected prompts by category sampled from the Parti Prompts (P2) dataset.}
\label{tab:selected-prompts}
\centering
\small
\setlength{\tabcolsep}{4pt}
\renewcommand{\arraystretch}{1.05}
\begin{tabularx}{\linewidth}{>{\raggedright\arraybackslash}p{0.24\linewidth} >{\raggedright\arraybackslash}X}
\hline
Category & Prompt \\
\hline
Abstract & happiness \\
Abstract & metal \\
Abstract & element \\
Vehicles & an airplane taking off of a runway \\
Vehicles & an antique car by a beach \\
Vehicles & a drop-top sports car coming around a bend in the road \\
Illustrations & a cube made of porcupine \\
Illustrations & a musical note \\
Illustrations & a sketch of a skyscraper \\
Arts & an oil surrealist painting of a dreamworld on a seashore where clocks and watches appear to be inexplicably limp and melting in the desolate landscape. a table on the left, with a golden watch swarmed by ants. a strange fleshy creature in the center of the painting \\
Arts & A raccoon wearing formal clothes, wearing a tophat and holding a cane. The raccoon is holding a garbage bag. Oil painting in the style of Vincent Van Gogh. \\
Arts & an abstract painting of a tree and a building \\
World Knowledge & The Statue of Liberty \\
World Knowledge & the grand canyon \\
World Knowledge & A portrait of a metal statue of a pharaoh wearing steampunk glasses and a leather jacket over a white t-shirt that has a drawing of a space shuttle on it. \\
People & a knight holding a long sword \\
People & an elder politician giving a campaign speech \\
People & children on a couch \\
Animals & Portrait of a tiger wearing a train conductor's hat and holding a skateboard that has a yin-yang symbol on it. woodcut \\
Animals & a cat \\
Animals & Portrait of a gecko wearing a train conductor’s hat and holding a flag that has a yin-yang symbol on it. Oil on canvas. \\
Artifacts & a yellow diamond-shaped sign with a deer silhouette \\
Artifacts & a black t-shirt with the peace sign on it \\
Artifacts & a yield sign \\
Food \& Beverage & a glass of orange juice with an orange peel stuck on the rim \\
Food \& Beverage & a roast turkey on the table \\
Food \& Beverage & A castle made of tortilla chips, in a river made of salsa. There are tiny burritos walking around the castle \\
Produce \& Plants & a walnut \\
Produce \& Plants & A photo of a palm tree made of water. \\
Produce \& Plants & a tree reflected in the hood of a blue car \\
Outdoor Scenes & a street with several cars on it \\
Outdoor Scenes & two chemtrails forming an X in blue sky \\
Outdoor Scenes & a house with no windows \\
Indoor Scenes & An empty fireplace with a television above it. The TV shows a lion hugging a giraffe. \\
Indoor Scenes & a wood cabin with a fire pit in front of it \\
Indoor Scenes & a bunch of laptops piled on a sofa \\
\hline
\end{tabularx}
\end{table}

\newpage

\section{Final Generated Images}

\begin{figure}[htb]
  \centering
  \includegraphics[width=7.7cm]{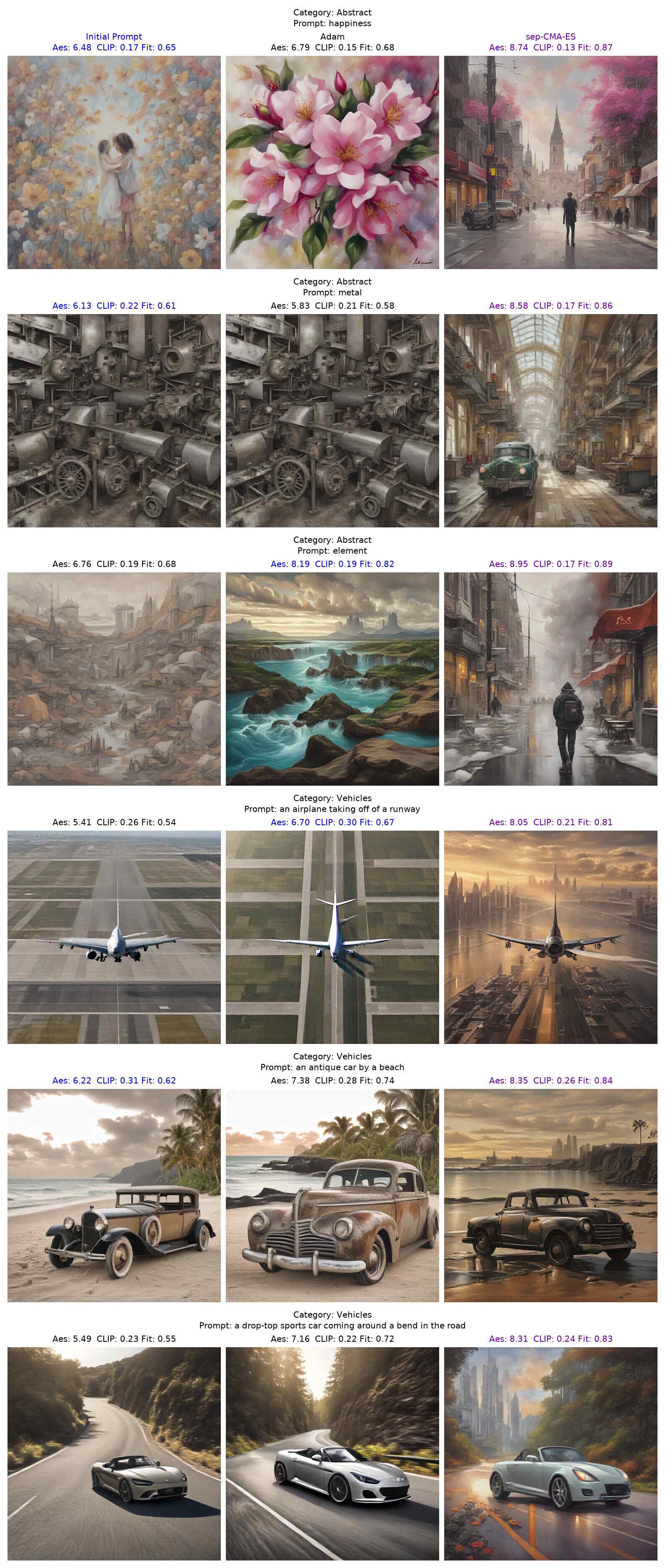}
  \caption{Final outputs from baseline SDXL Turbo, Adam, and sep-CMA-ES for prompts 1 to 6 in the aesthetics-only setting. Rows correspond to prompts and columns to methods, with aesthetic, CLIP, and fitness scores above each image; purple marks the highest-fitness image, while red or blue mark the best aesthetic or CLIPScore when they do not match the fitness optimum.}
  
  \label{fig:sample_results_grid_a100_b0_1}
\end{figure}

\begin{figure}[htb]
  \centering
  \includegraphics[width=7.7cm]{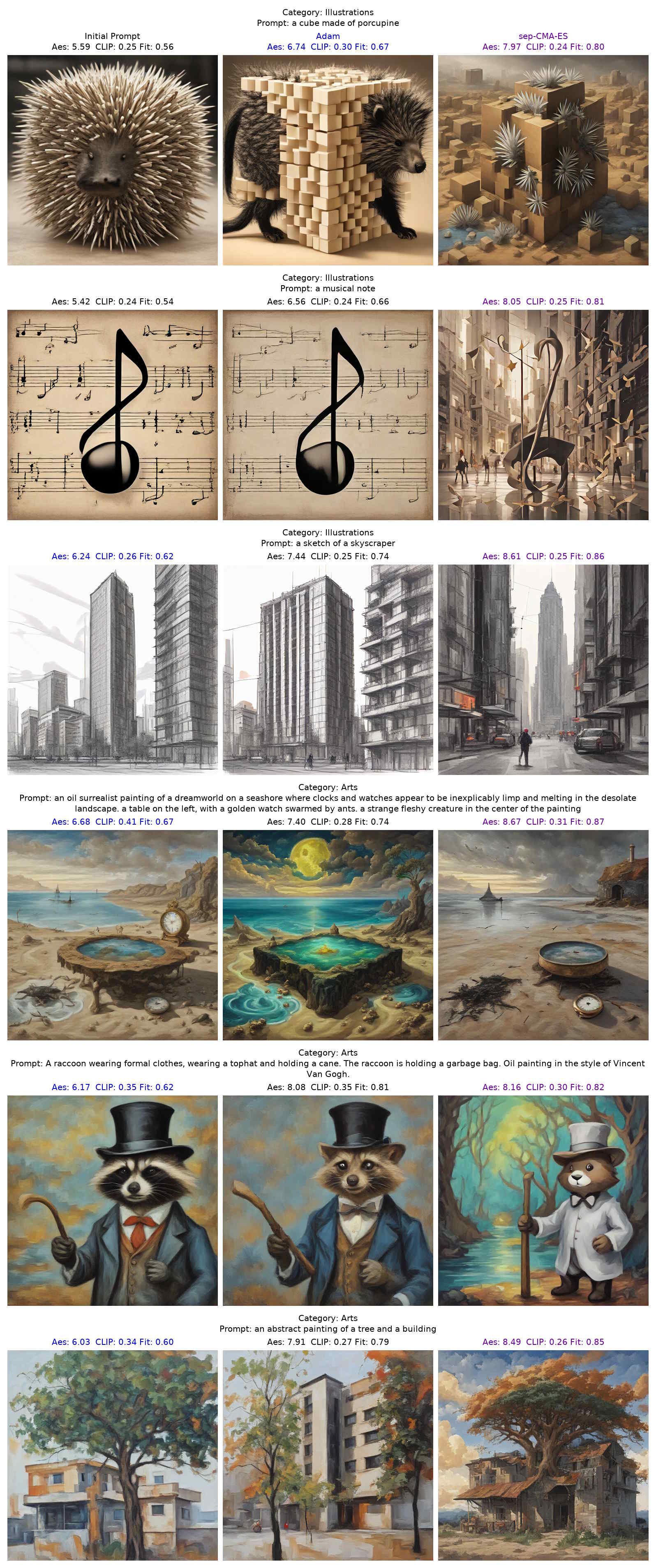}
  \caption{Final outputs from baseline SDXL Turbo, Adam, and sep-CMA-ES for prompts 7 to 12 in the aesthetics-only setting. Rows correspond to prompts and columns to methods, with aesthetic, CLIP, and fitness scores above each image; purple marks the highest-fitness image, while red or blue mark the best aesthetic or CLIPScore when they do not match the fitness optimum.}
  
  \label{fig:sample_results_grid_a100_b0_2}
\end{figure}

\begin{figure}[htb]
  \centering
  \includegraphics[width=7.7cm]{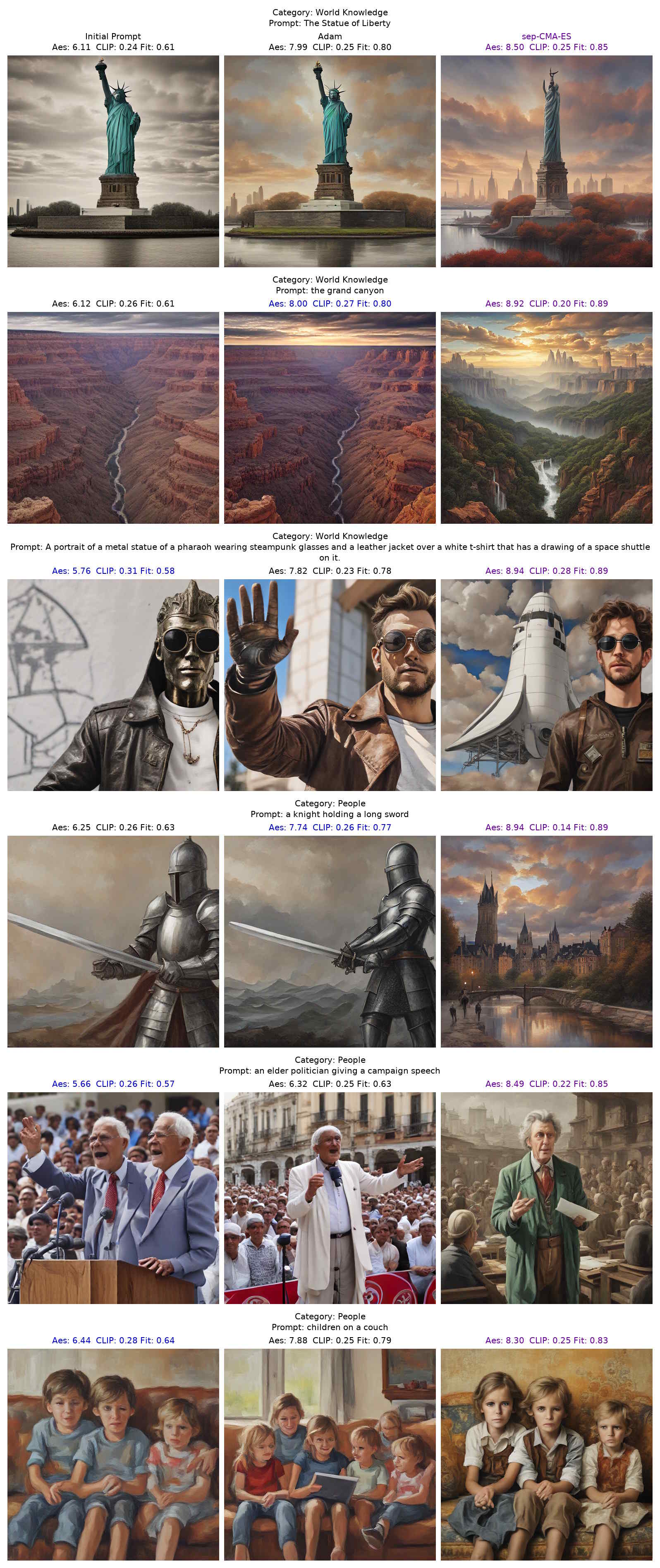}
  \caption{Final outputs from baseline SDXL Turbo, Adam, and sep-CMA-ES for prompts 13 to 18 in the aesthetics-only setting. Rows correspond to prompts and columns to methods, with aesthetic, CLIP, and fitness scores above each image; purple marks the highest-fitness image, while red or blue mark the best aesthetic or CLIPScore when they do not match the fitness optimum.}
  
  \label{fig:sample_results_grid_a100_b0_3}
\end{figure}

\begin{figure}[htb]
  \centering
  \includegraphics[width=7.7cm]{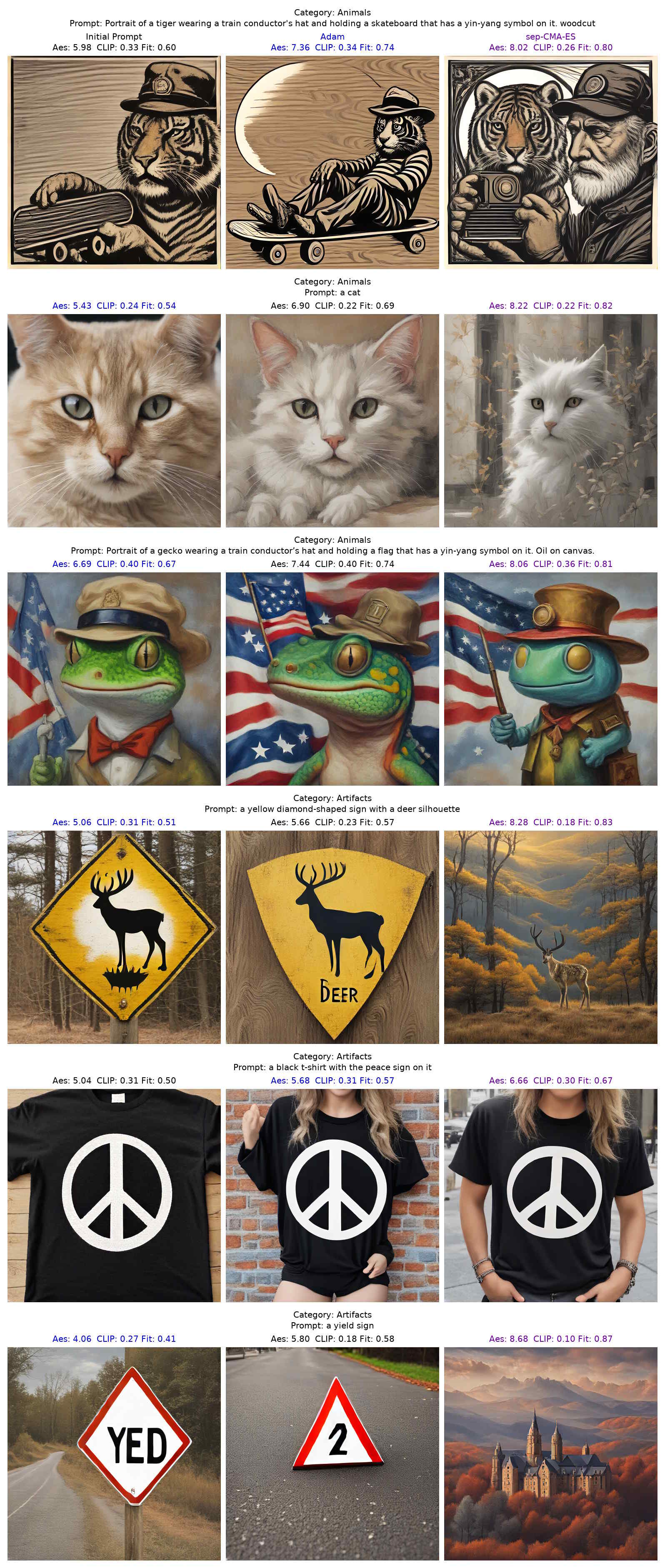}
  \caption{Final outputs from baseline SDXL Turbo, Adam, and sep-CMA-ES for prompts 19 to 24 in the aesthetics-only setting. Rows correspond to prompts and columns to methods, with aesthetic, CLIP, and fitness scores above each image; purple marks the highest-fitness image, while red or blue mark the best aesthetic or CLIPScore when they do not match the fitness optimum.}
  
  \label{fig:sample_results_grid_a100_b0_4}
\end{figure}

\begin{figure}[htb]
  \centering
  \includegraphics[width=7.7cm]{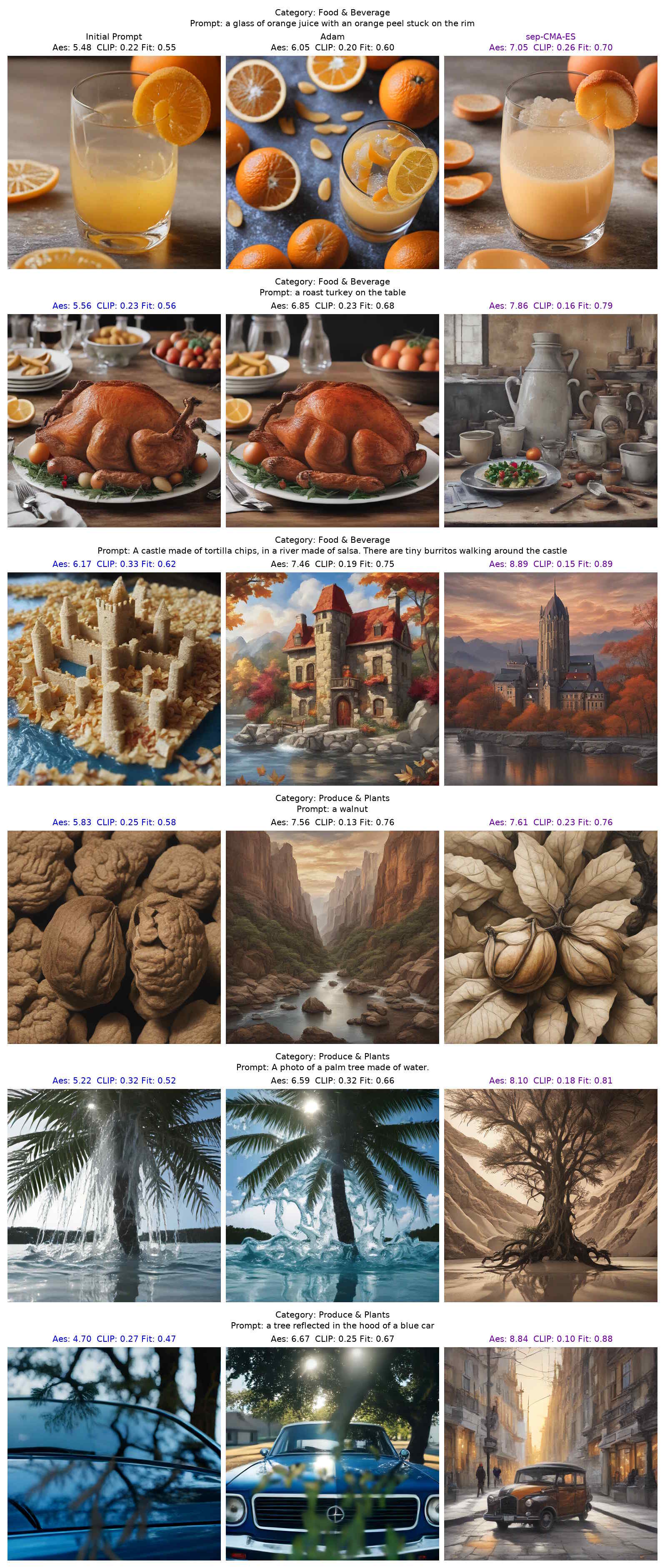}
  \caption{Final outputs from baseline SDXL Turbo, Adam, and sep-CMA-ES for prompts 25 to 30 in the aesthetics-only setting. Rows correspond to prompts and columns to methods, with aesthetic, CLIP, and fitness scores above each image; purple marks the highest-fitness image, while red or blue mark the best aesthetic or CLIPScore when they do not match the fitness optimum.}
  
  \label{fig:sample_results_grid_a100_b0_5}
\end{figure}

\begin{figure}[htb]
  \centering
  \includegraphics[width=7.7cm]{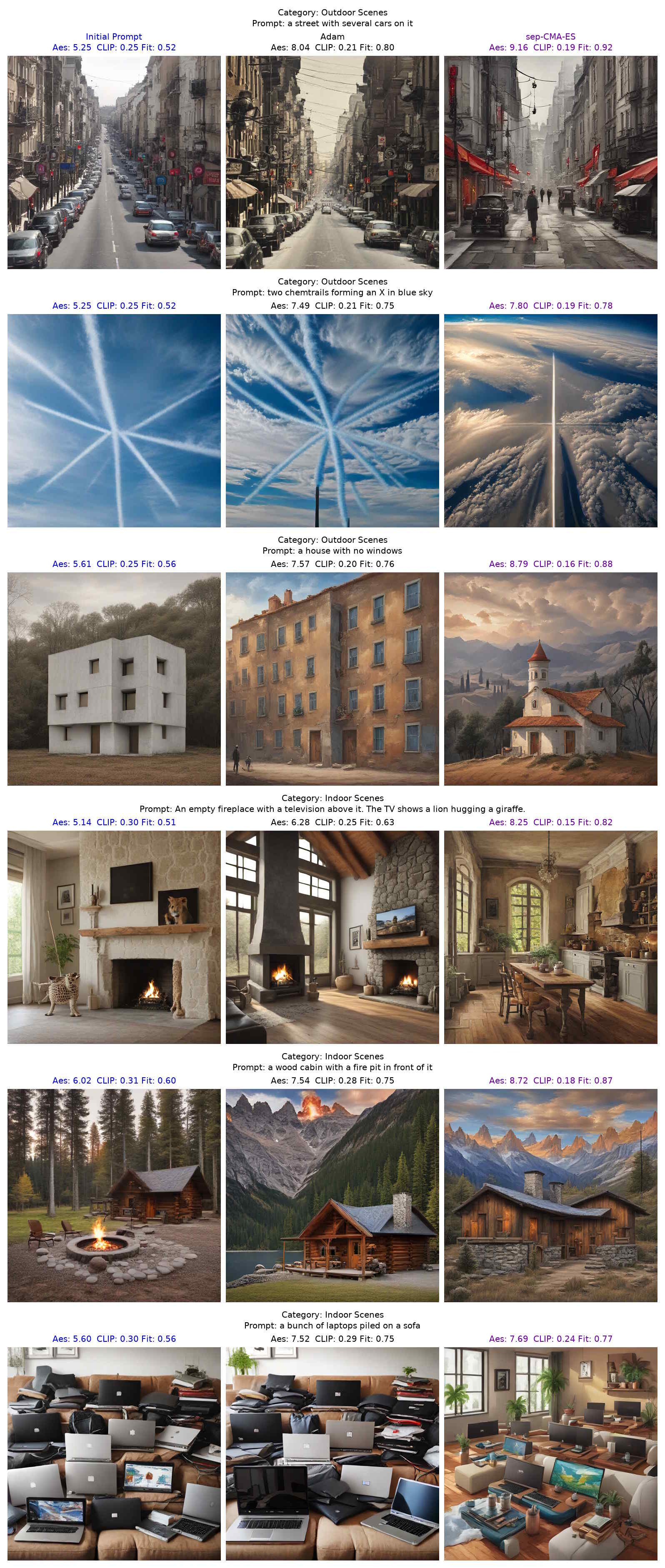}
  \caption{Final outputs from baseline SDXL Turbo, Adam, and sep-CMA-ES for prompts 31 to 36 in the aesthetics-only setting. Rows correspond to prompts and columns to methods, with aesthetic, CLIP, and fitness scores above each image; purple marks the highest-fitness image, while red or blue mark the best aesthetic or CLIPScore when they do not match the fitness optimum.}
  
  \label{fig:sample_results_grid_a100_b0_6}
\end{figure}

\begin{figure}[htb]
  \centering
  \includegraphics[width=7.7cm]{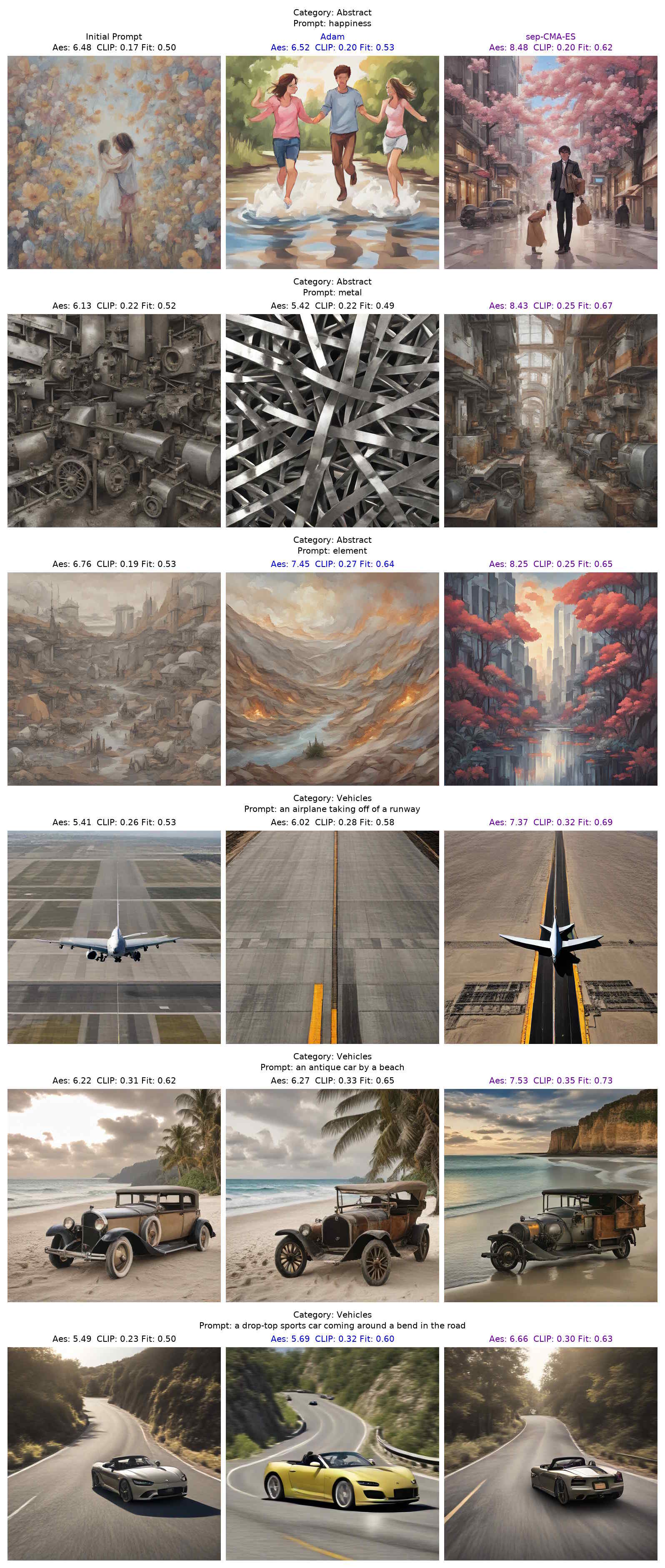}
  \caption{Final outputs from baseline SDXL Turbo, Adam, and sep-CMA-ES for prompts 1 to 6 in the balanced setting. Rows correspond to prompts and columns to methods, with aesthetic, CLIP, and fitness scores above each image; purple marks the highest-fitness image, while red or blue mark the best aesthetic or CLIPScore when they do not match the fitness optimum.}
  
  \label{fig:sample_results_grid_a50_b50_1}
\end{figure}

\begin{figure}[htb]
  \centering
  \includegraphics[width=7.7cm]{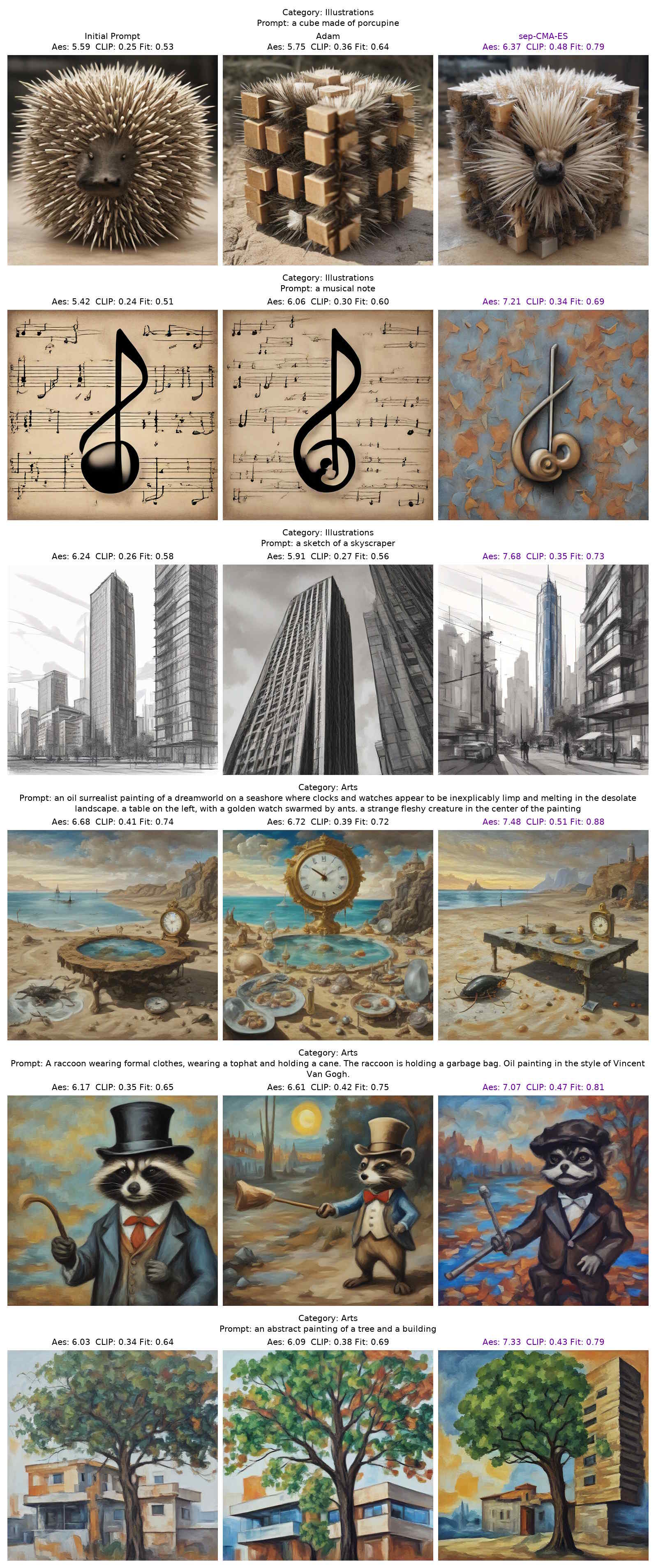}
  \caption{Final outputs from baseline SDXL Turbo, Adam, and sep-CMA-ES for prompts 7 to 12 in the balanced setting. Rows correspond to prompts and columns to methods, with aesthetic, CLIP, and fitness scores above each image; purple marks the highest-fitness image, while red or blue mark the best aesthetic or CLIPScore when they do not match the fitness optimum.}
  
  \label{fig:sample_results_grid_a50_b50_2}
\end{figure}

\begin{figure}[htb]
  \centering
  \includegraphics[width=7.7cm]{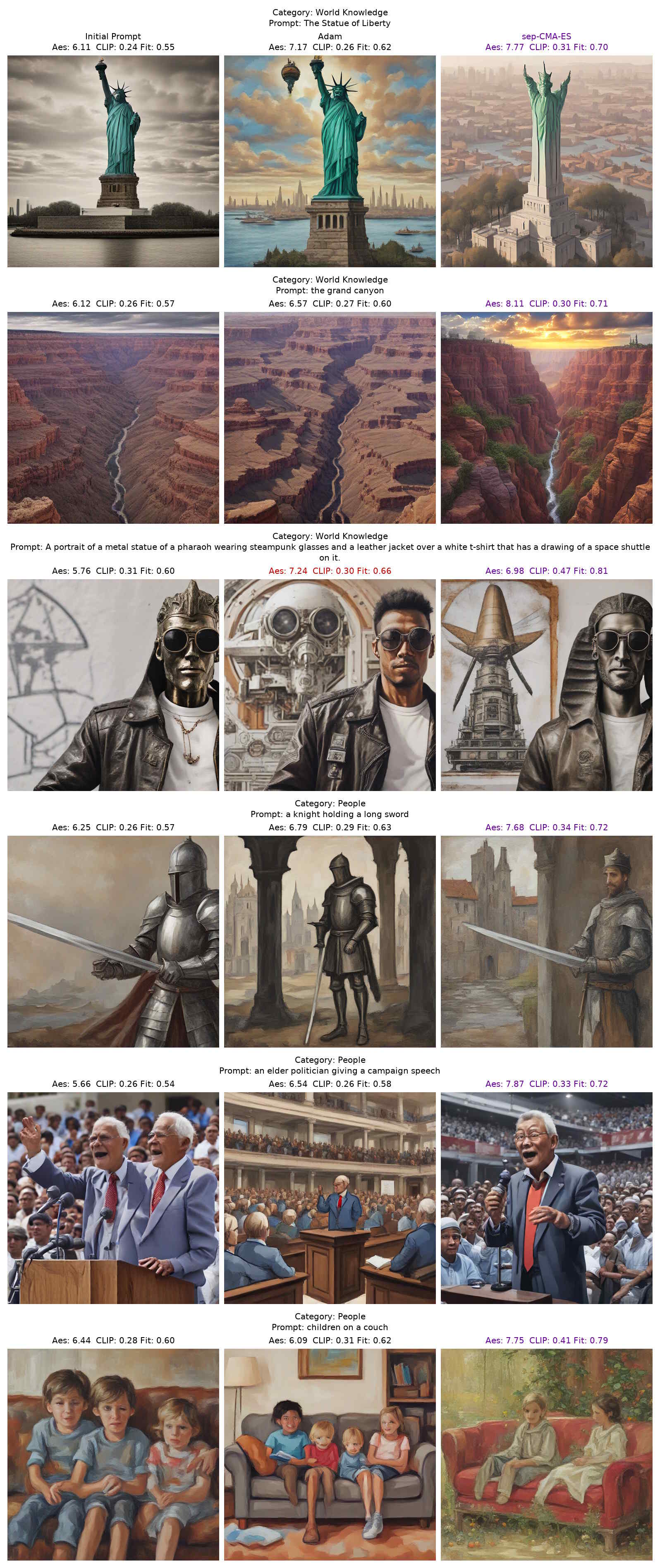}
  \caption{Final outputs from baseline SDXL Turbo, Adam, and sep-CMA-ES for prompts 13 to 18 in the balanced setting. Rows correspond to prompts and columns to methods, with aesthetic, CLIP, and fitness scores above each image; purple marks the highest-fitness image, while red or blue mark the best aesthetic or CLIPScore when they do not match the fitness optimum.}
  
  \label{fig:sample_results_grid_a50_b50_3}
\end{figure}

\begin{figure}[htb]
  \centering
  \includegraphics[width=7.7cm]{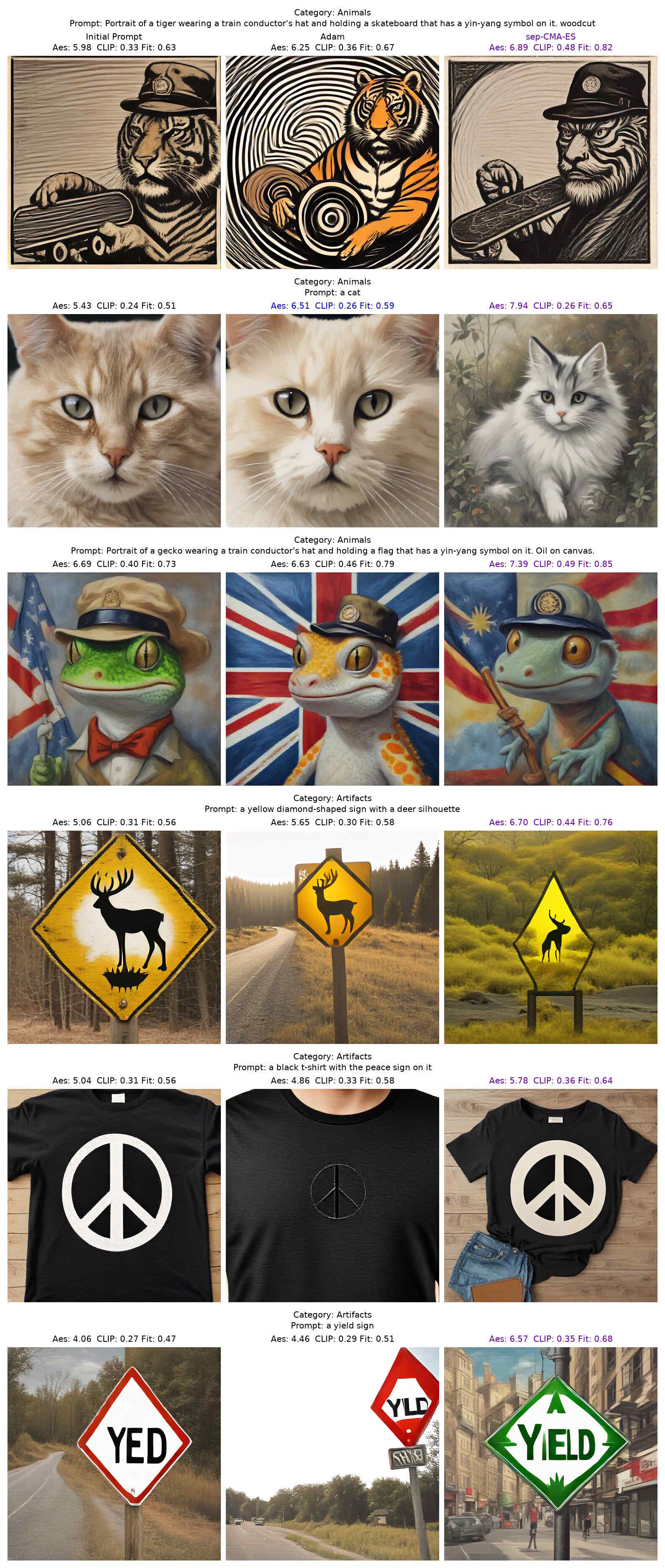}
  \caption{Final outputs from baseline SDXL Turbo, Adam, and sep-CMA-ES for prompts 19 to 24 in the balanced setting. Rows correspond to prompts and columns to methods, with aesthetic, CLIP, and fitness scores above each image; purple marks the highest-fitness image, while red or blue mark the best aesthetic or CLIPScore when they do not match the fitness optimum.}
  
  \label{fig:sample_results_grid_a50_b50_4}
\end{figure}

\begin{figure}[htb]
  \centering
  \includegraphics[width=7.7cm]{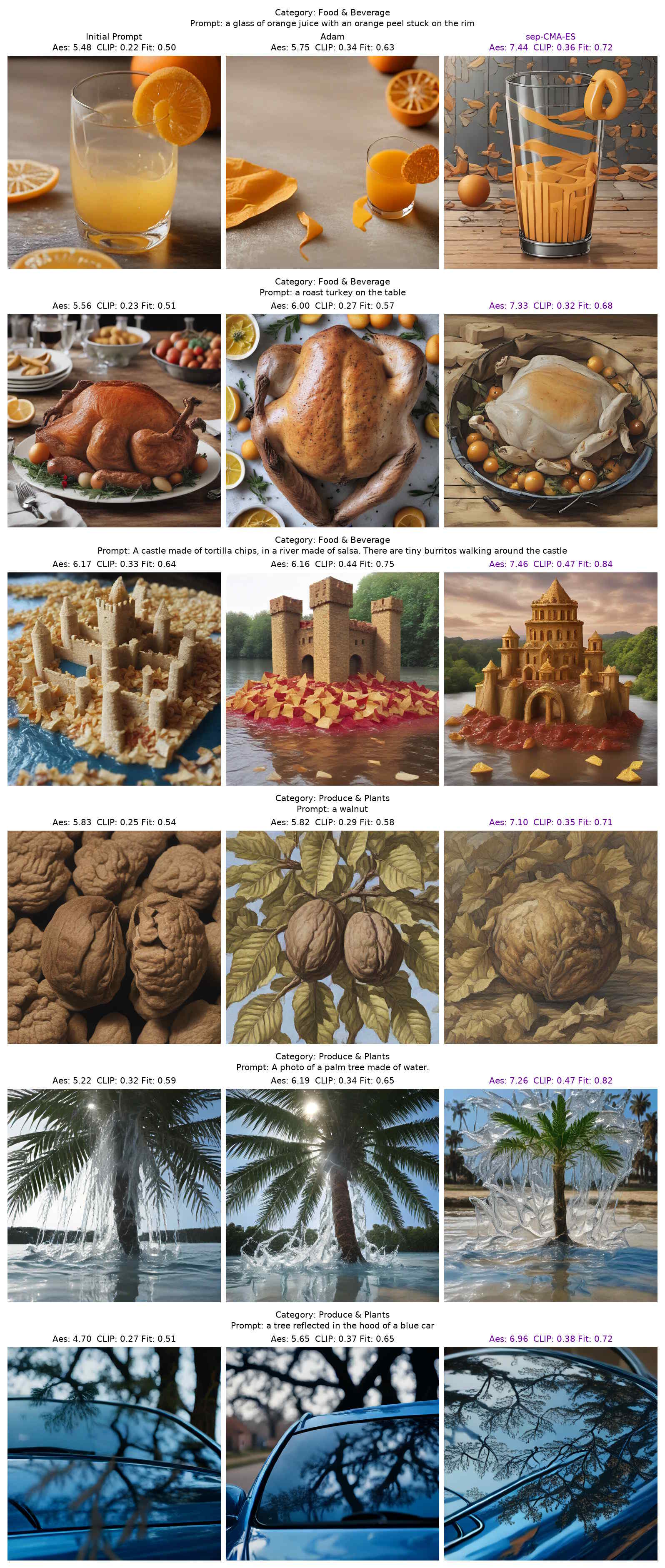}
  \caption{Final outputs from baseline SDXL Turbo, Adam, and sep-CMA-ES for prompts 25 to 30 in the balanced setting. Rows correspond to prompts and columns to methods, with aesthetic, CLIP, and fitness scores above each image; purple marks the highest-fitness image, while red or blue mark the best aesthetic or CLIPScore when they do not match the fitness optimum.}
  
  \label{fig:sample_results_grid_a50_b50_5}
\end{figure}

\begin{figure}[htb]
  \centering
  \includegraphics[width=7.7cm]{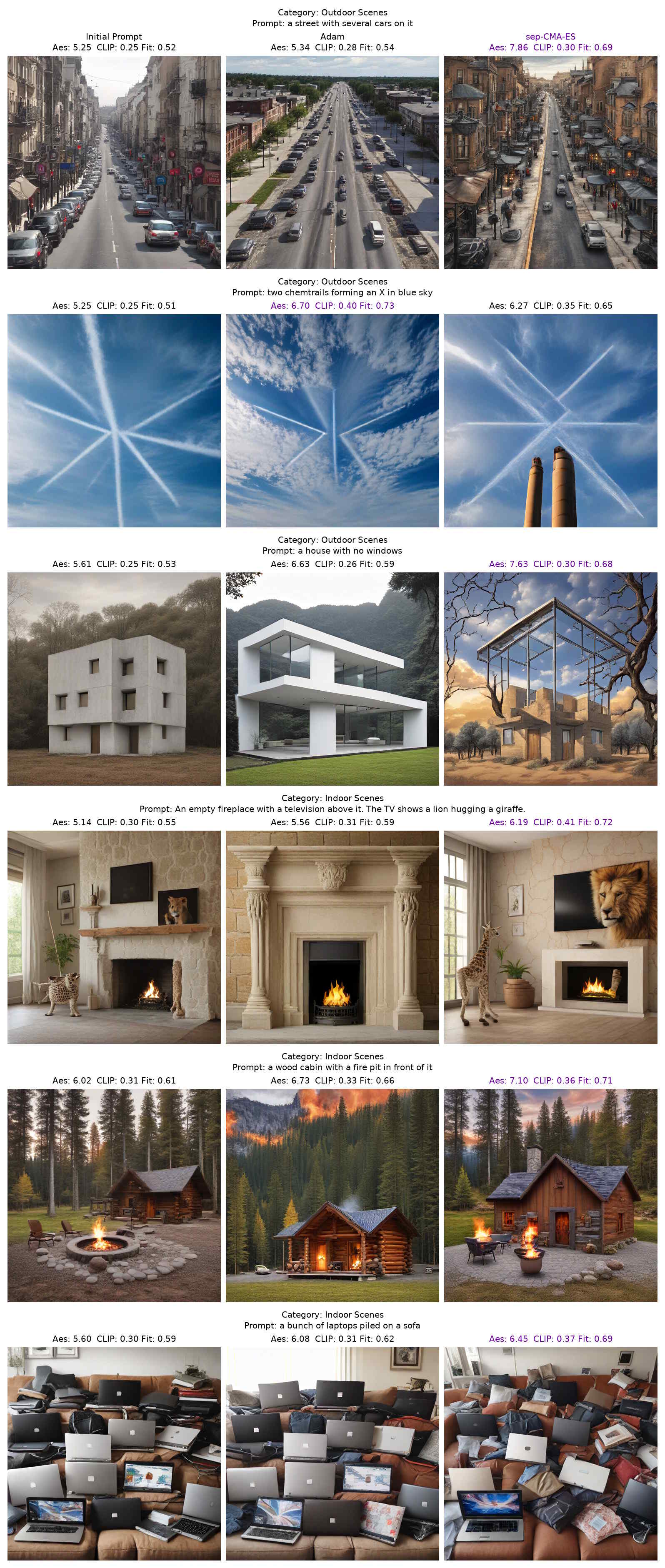}
  \caption{Final outputs from baseline SDXL Turbo, Adam, and sep-CMA-ES for prompts 31 to 36 in the balanced setting. Rows correspond to prompts and columns to methods, with aesthetic, CLIP, and fitness scores above each image; purple marks the highest-fitness image, while red or blue mark the best aesthetic or CLIPScore when they do not match the fitness optimum.}
  
  \label{fig:sample_results_grid_a50_b50_6}
\end{figure}

\begin{figure}[htb]
  \centering
  \includegraphics[width=7.7cm]{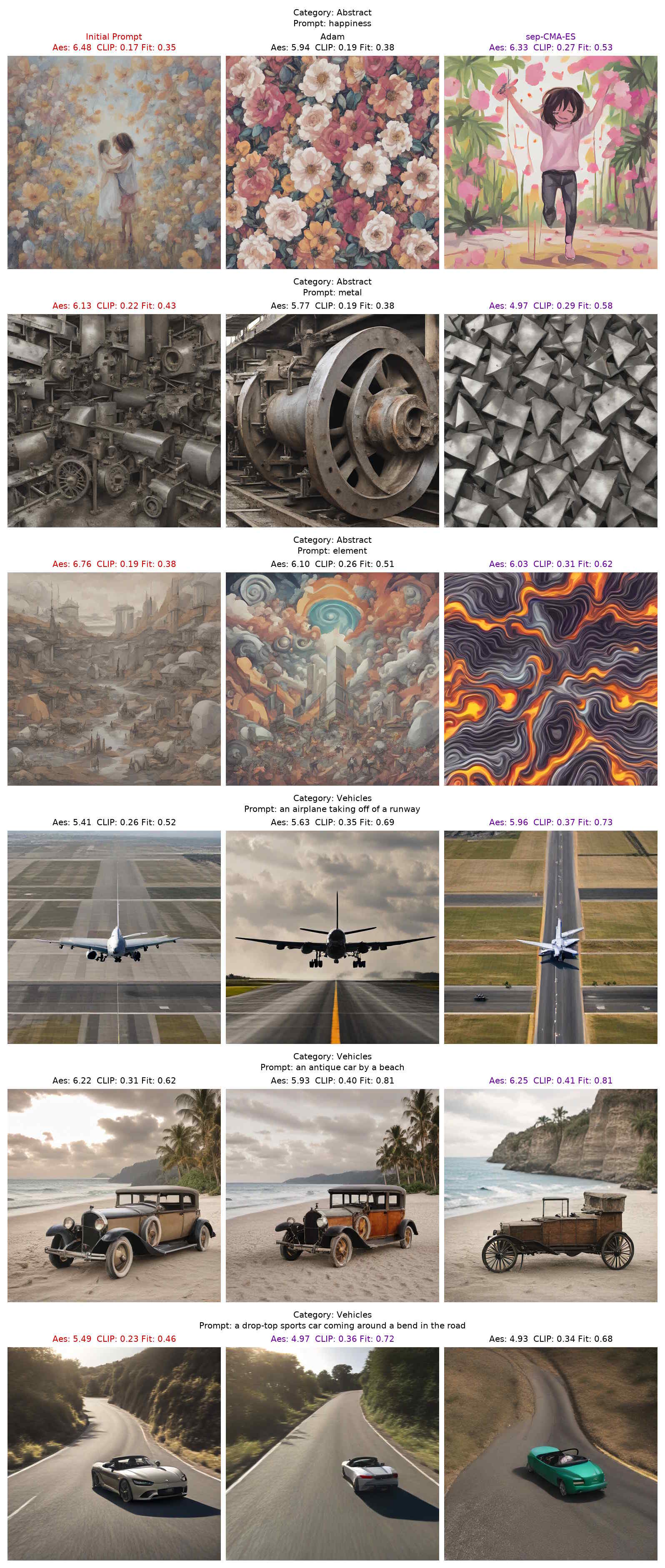}
  \caption{Final outputs from baseline SDXL Turbo, Adam, and sep-CMA-ES for prompts 1 to 6 in the prompt-image alignment only setting. Rows correspond to prompts and columns to methods, with aesthetic, CLIP, and fitness scores above each image; purple marks the highest-fitness image, while red or blue mark the best aesthetic or CLIPScore when they do not match the fitness optimum.}
  
  \label{fig:sample_results_grid_a0_b100_1}
\end{figure}

\begin{figure}[htb]
  \centering
  \includegraphics[width=7.7cm]{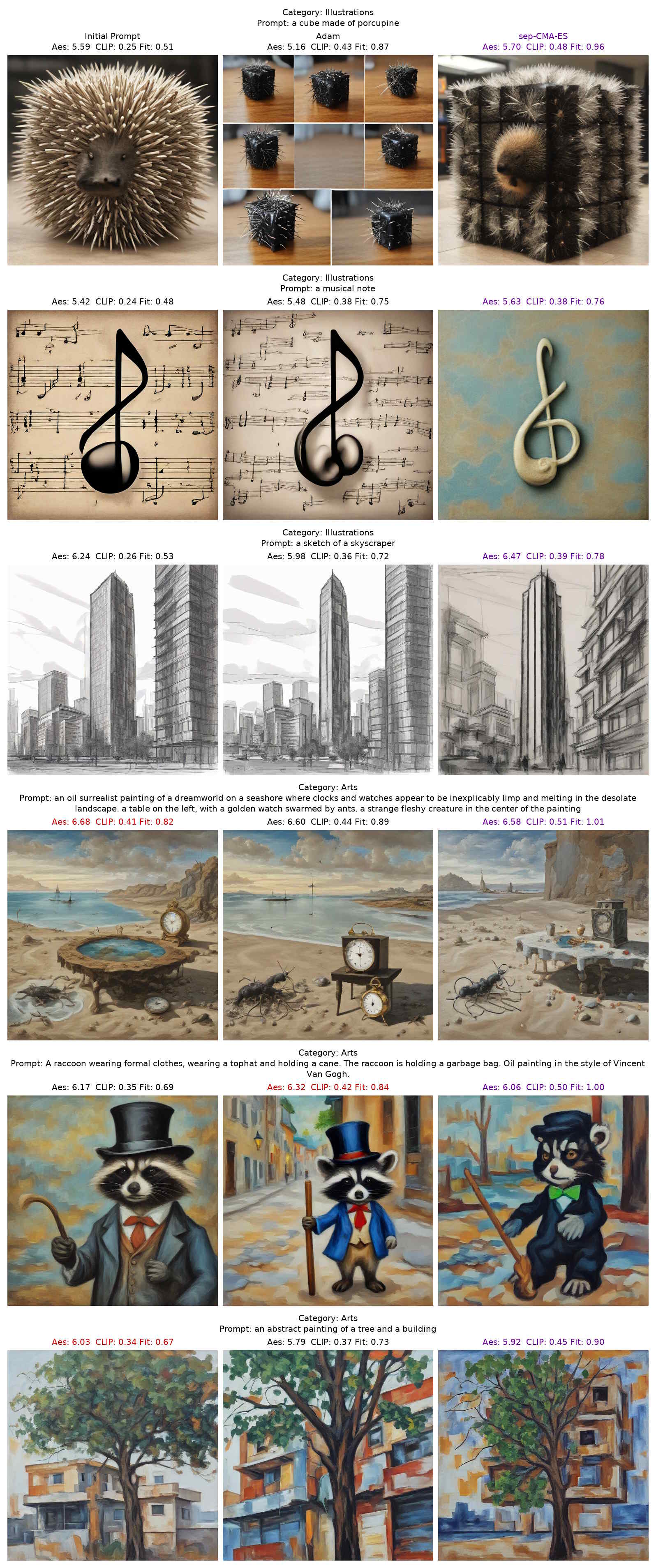}
  \caption{Final outputs from baseline SDXL Turbo, Adam, and sep-CMA-ES for prompts 7 to 12 in the prompt-image alignment only setting. Rows correspond to prompts and columns to methods, with aesthetic, CLIP, and fitness scores above each image; purple marks the highest-fitness image, while red or blue mark the best aesthetic or CLIPScore when they do not match the fitness optimum.}
  
  \label{fig:sample_results_grid_a0_b100_2}
\end{figure}

\begin{figure}[htb]
  \centering
  \includegraphics[width=7.7cm]{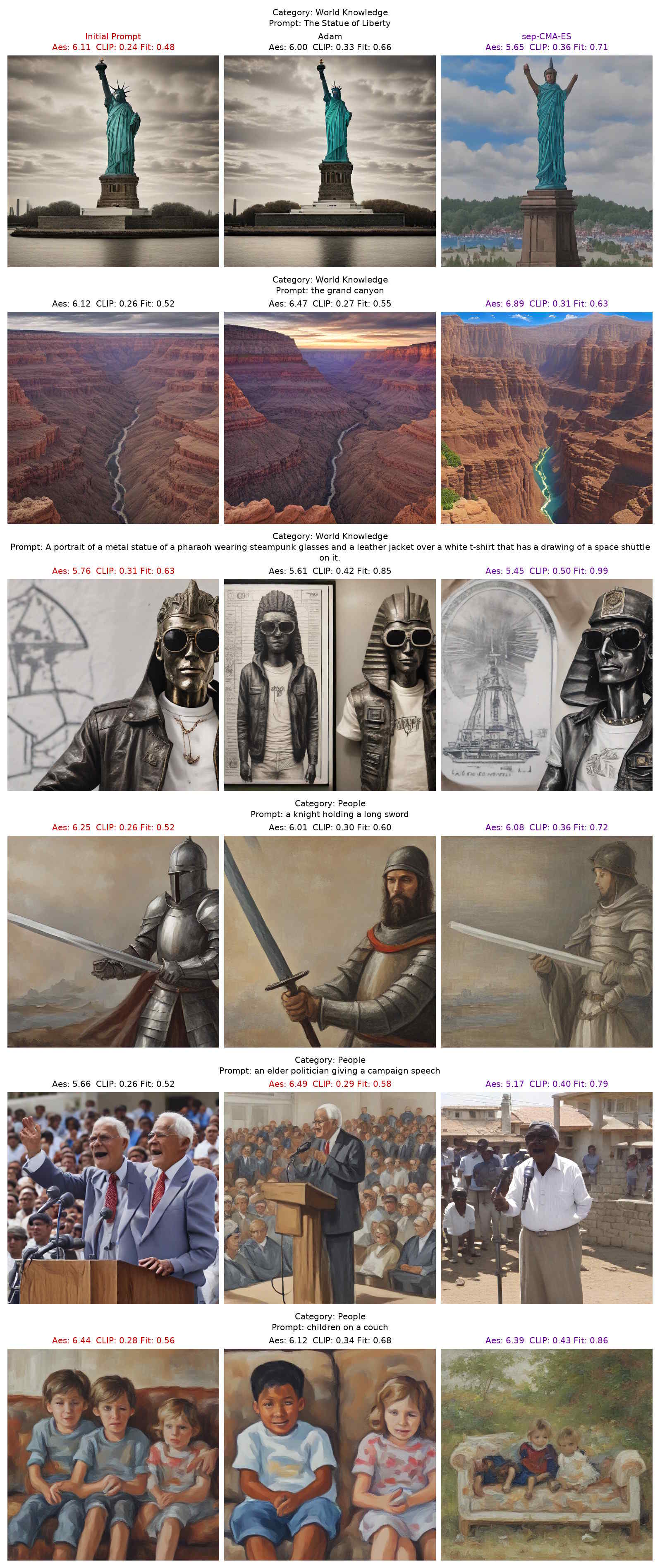}
  \caption{Final outputs from baseline SDXL Turbo, Adam, and sep-CMA-ES for prompts 13 to 18 in the prompt-image alignment only setting. Rows correspond to prompts and columns to methods, with aesthetic, CLIP, and fitness scores above each image; purple marks the highest-fitness image, while red or blue mark the best aesthetic or CLIPScore when they do not match the fitness optimum.}
  
  \label{fig:sample_results_grid_a0_b100_3}
\end{figure}

\begin{figure}[htb]
  \centering
  \includegraphics[width=7.7cm]{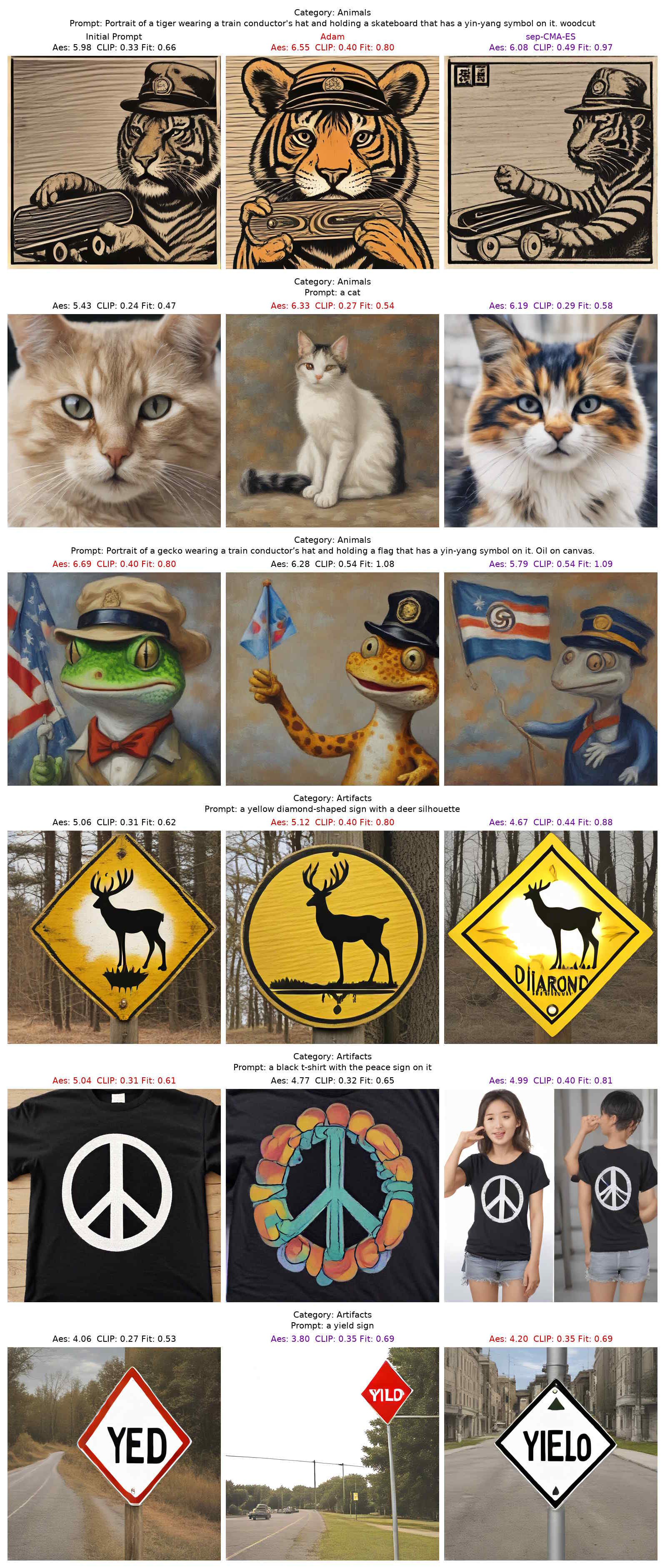}
  \caption{Final outputs from baseline SDXL Turbo, Adam, and sep-CMA-ES for prompts 19 to 24 in the prompt-image alignment only setting. Rows correspond to prompts and columns to methods, with aesthetic, CLIP, and fitness scores above each image; purple marks the highest-fitness image, while red or blue mark the best aesthetic or CLIPScore when they do not match the fitness optimum.}
  
  \label{fig:sample_results_grid_a0_b100_4}
\end{figure}

\begin{figure}[htb]
  \centering
  \includegraphics[width=7.7cm]{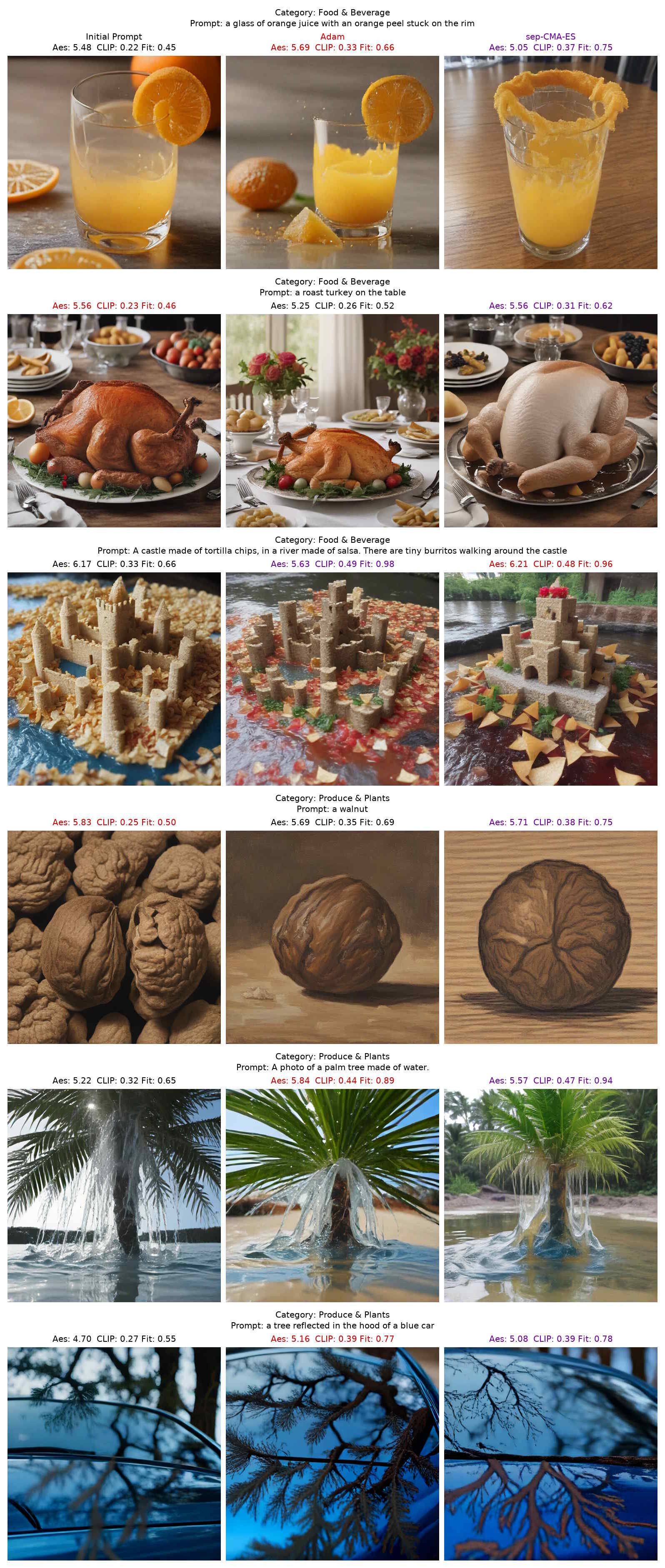}
  \caption{Final outputs from baseline SDXL Turbo, Adam, and sep-CMA-ES for prompts 25 to 30 in the prompt-image alignment only setting. Rows correspond to prompts and columns to methods, with aesthetic, CLIP, and fitness scores above each image; purple marks the highest-fitness image, while red or blue mark the best aesthetic or CLIPScore when they do not match the fitness optimum.}
  
  \label{fig:sample_results_grid_a0_b100_5}
\end{figure}

\begin{figure}[htb]
  \centering
  \includegraphics[width=7.7cm]{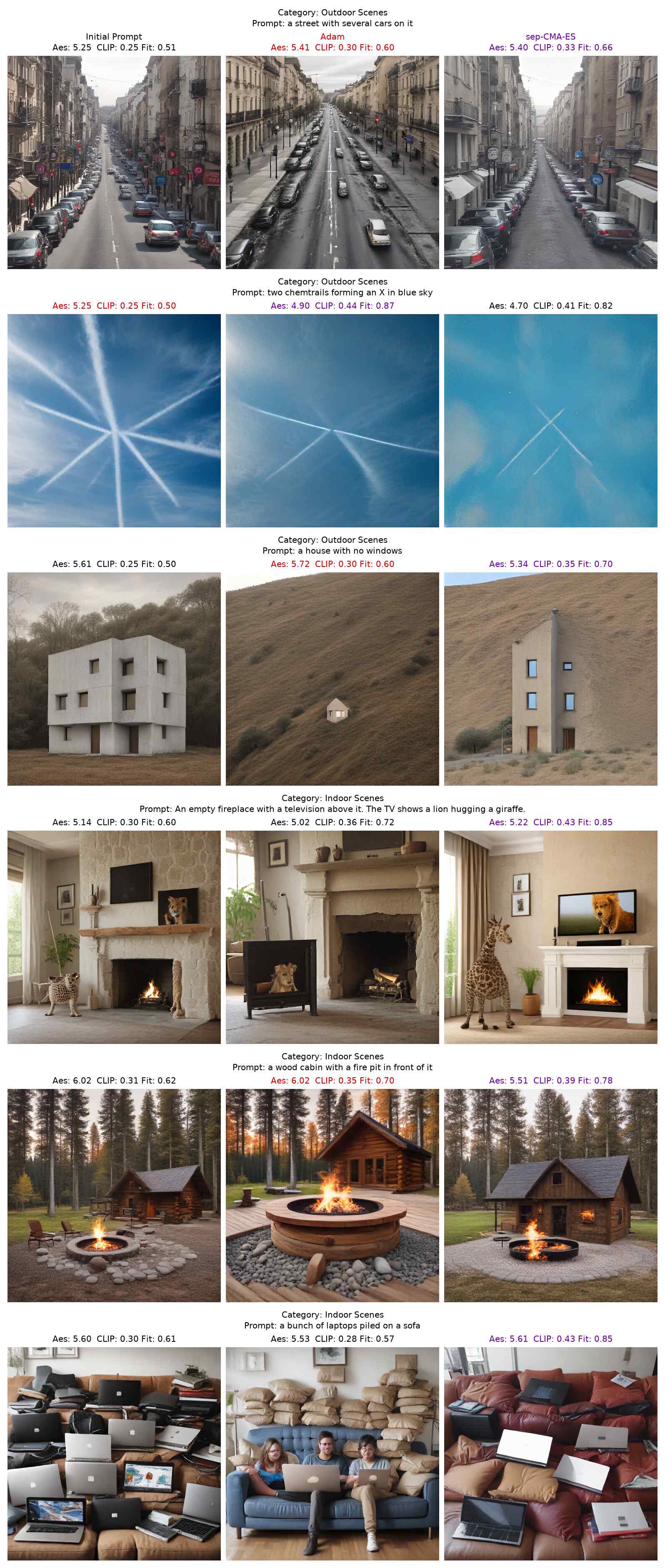}
  \caption{Final outputs from baseline SDXL Turbo, Adam, and sep-CMA-ES for prompts 31 to 36 in the prompt-image alignment only setting. Rows correspond to prompts and columns to methods, with aesthetic, CLIP, and fitness scores above each image; purple marks the highest-fitness image, while red or blue mark the best aesthetic or CLIPScore when they do not match the fitness optimum.}
  
  \label{fig:sample_results_grid_a0_b100_6}
\end{figure}

\end{document}